\documentclass[10pt,twocolumn,letterpaper]{article}
\usepackage{abstract}
\usepackage{caption}
\usepackage{subcaption}
\usepackage{cvpr}              

\usepackage[accsupp]{axessibility} 
\usepackage{graphicx}
\usepackage{amsmath}
\usepackage{amssymb}
\usepackage{booktabs}

\usepackage{algorithm}
\usepackage{algpseudocode}
\usepackage{adjustbox}
\usepackage[table,xcdraw]{xcolor}
\usepackage[flushleft]{threeparttable}
\usepackage{makecell}
\usepackage{framed,multirow}
\usepackage[subtle]{savetrees}

\newtheorem{definition}{Definition}

\makeatletter
\renewcommand{\paragraph}{%
  \@startsection{paragraph}{4}%
  {\z@}{1ex \@plus 1ex \@minus .2ex}{-1em}%
  {\normalfont\normalsize\bfseries}%
}
\makeatother

\usepackage[bookmarks=false]{hyperref}
\hypersetup{pagebackref,breaklinks,colorlinks}

\usepackage[capitalize]{cleveref}
\crefname{section}{Sec.}{Secs.}
\Crefname{section}{Section}{Sections}
\Crefname{table}{Table}{Tables}
\crefname{table}{Tab.}{Tabs.}
\Crefname{algorithm}{Algorithm}{Algorithms}

\def\cvprPaperID{3592} 
\def\confName{CVPR}
\def\confYear{2023}

\newcommand{\methodname}{\mbox{MinD-Vis}} 

\begin{document}

\title{Seeing Beyond the Brain: 
Conditional Diffusion Model with\\ Sparse Masked Modeling for Vision Decoding}

\author{
Zijiao~Chen\textsuperscript{1}\thanks{Equal contributions.} \and
\newcommand\CoAuthorMark{\footnotemark[\arabic{footnote}]}
Jiaxin~Qing\textsuperscript{2}\CoAuthorMark \and
Tiange~Xiang\textsuperscript{3} \and
Wan~Lin~Yue\textsuperscript{1} \and
Juan~Helen~Zhou\textsuperscript{1}\thanks{Corresponding author (\href{mailto:helen.zhou@nus.edu.sg}{helen.zhou@nus.edu.sg})}
}
\date{
\normalsize
\textsuperscript{1}National University of Singapore,
\textsuperscript{2}The Chinese University of Hong Kong, 
\textsuperscript{3}Stanford University \\
\url{https://mind-vis.github.io}
}

\twocolumn[{%
\renewcommand\twocolumn[1]{#1}%
\maketitle
\begin{center}
  \captionsetup{type=figure}
  \setlength{\abovecaptionskip}{-1pt}
  \vspace{-3em}
\includegraphics[width=0.95\textwidth]{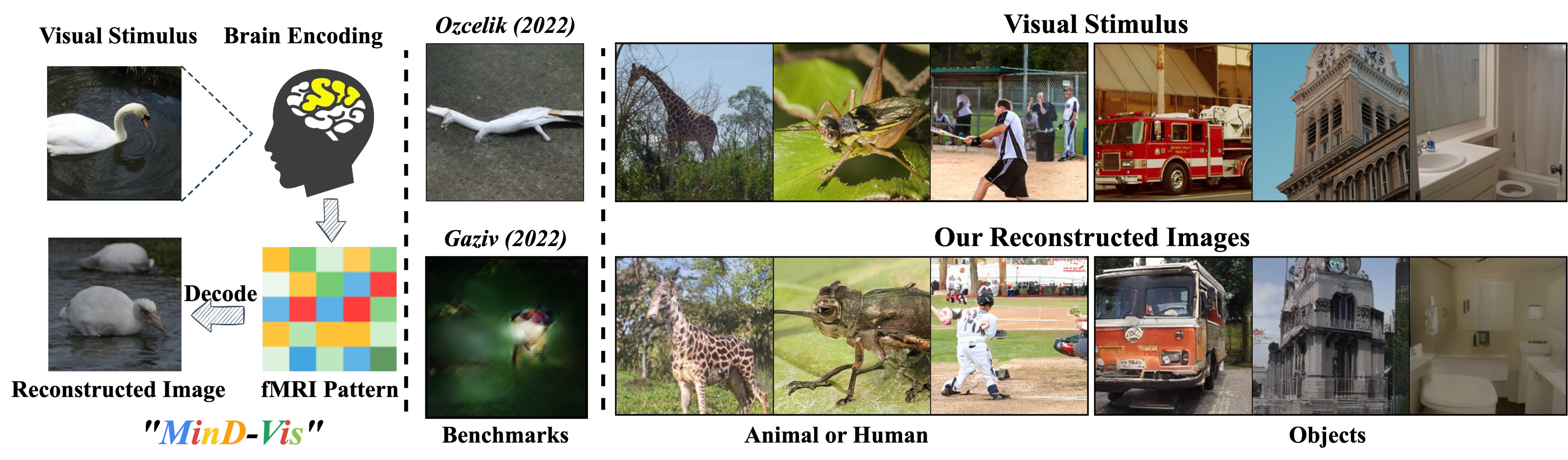}
    \caption{\textbf{Brain Decoding and Image Reconstruction}. For the first time, our proposed \textbf{\methodname} is capable of decoding fMRI-based brain activities and reconstructing images with not only plausible details but also accurate semantics and image features (texture, shape, \etc), pushing this domain a considerable step forward. Left: Task overview. Middle: Comparison with benchmarks. Right: More reconstruction examples. }
    \label{fig:intro}
    
\end{center}
}]
\saythanks

\begin{abstract}
Decoding visual stimuli from brain recordings aims to deepen our understanding of the human visual system and build a solid foundation for bridging human and computer vision through the Brain-Computer Interface.
However, reconstructing high-quality images with correct semantics from brain recordings is a challenging problem due to the complex underlying representations of brain signals and the scarcity of data annotations.
In this work, we present \textbf{\methodname}: Sparse \textbf{M}asked Bra\textbf{in} Modeling with Double-Conditioned Latent \textbf{D}iffusion Model for Human \textbf{Vis}ion Decoding.
Firstly, we learn an effective self-supervised representation of fMRI data using mask modeling in a large latent space inspired by the sparse coding of information in the primary visual cortex.
Then by augmenting a latent diffusion model with double-conditioning,
we show that {\methodname} can reconstruct highly plausible images with semantically matching details from brain recordings using very few paired annotations. 
We benchmarked our model qualitatively and quantitatively; the experimental results indicate that our method outperformed state-of-the-art in both semantic mapping (100-way semantic classification) and generation quality (FID) by \textbf{$66\%$} and \textbf{$41\%$} respectively. 
An exhaustive ablation study was also conducted to analyze our framework. 

\end{abstract}

\section{Introduction}
\label{sec:intro}
\emph{``What you think is what you see''}. 
Human perception and prior knowledge are deeply intertwined in one's mind~\cite{vis_mem}. Our perception of the world is determined not only by objective stimuli properties but also by our experiences, forming complex brain activities underlying our perception.
Understanding these brain activities and recovering the encoded information is a key goal in cognitive neuroscience. Within this broad objective, decoding visual information is one of the challenging problems that are the focus of a large body of literature~\cite{spike,mouse_spike,rgc_nips,cal_img1}.

As a non-invasive and effective method to measure brain activities indirectly, functional Magnetic Resonance Imaging (fMRI) is usually used to recover visual information, such as the image classes~\cite{kam2017,fmri2018}.   
With the help of recent deep learning models, it is intriguing if the original visual stimuli can be directly recovered from corresponding fMRI~\cite{shen2019,guy2019}, 
especially with the guidance of biological principles~\cite{tschopp2018connectome,sandin2020synaptic}.
However, due to the lack of fMRI-image pairs and useful biological guidance when decoding complex neural activity from fMRI directly, reconstructed images are usually blurry and semantically meaningless.
Thus it is crucial to learn effective and biological-valid representations for fMRI so that a clear and generalizable connection between brain activities and visual stimuli can be established with a few paired annotations. 

\begin{figure}[t]
\setlength{\abovecaptionskip}{-1pt}
\setlength{\belowcaptionskip}{-15pt}
  \centering
    \includegraphics[width=0.8\linewidth]{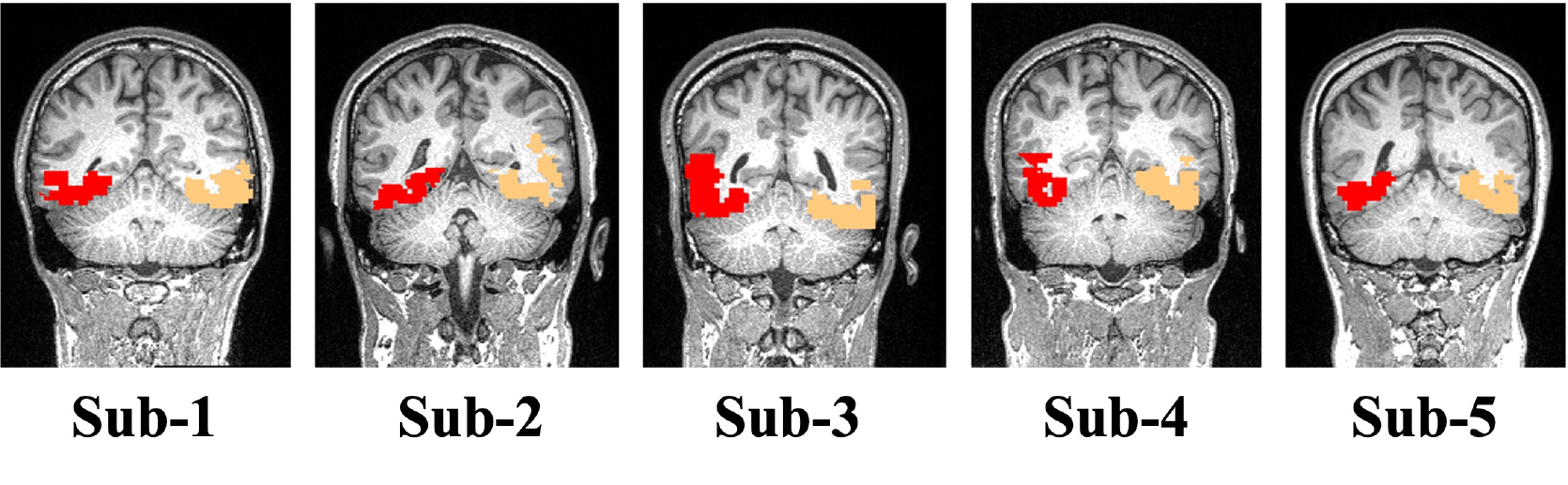}
    \caption{\textbf{Individual Differences in Regions Responding to Visual Stimuli.}
    Masks of the regions of interest activating during the same visual task differ in location and size across subjects. The primary visual cortex at the left (\textcolor{red}{red}) and the right (\textcolor{orange}{orange}) hemisphere are shown.}
    \label{fig:individual_diff}
\end{figure}

Moreover, individual variability in brain representations further complicates this problem. 
Individuals have unique brain activation patterns responding to the same visual stimulus (See~\cref{fig:individual_diff}). 
From the perspective of fMRI representation learning, a powerful brain decoding algorithm should robustly recognize features shared across the population over a background of individual variation~\cite{kam2017,chang2019bold5000}. 
On the other hand, we should also expect decoding variances due to the variation in individual perceptions. 
Therefore, we aim to learn representations from a large-scale dataset with rich demographic compositions and relax the direct generation from fMRI to conditional synthesis allowing for sampling variance under the same semantic category. 

Self-supervised learning with pretext tasks in large datasets is a powerful paradigm to distill the model with context knowledge.
A domain-specific downstream task (\eg classification) is usually adopted to finetune the pre-trained model further~\cite{clip,piti}, especially when the downstream dataset is small. 
Various pretext tasks are designed to benefit downstream tasks~\cite{vae, colorful}. 
Among these methods, Masked Signal Modeling (MSM) has achieved promising results in both vision~\cite{maeHe,simmim} and language understanding~\cite{bert,gpt} recently. 
At the same time, the probabilistic diffusion denoising model has shown its superior performance in content generation and training stability~\cite{openaiDiff}. 
A strong generation ability is also desired in our task to decode faithful visual stimuli from various categories.

Driven by the above analysis,
we propose \textbf{\methodname}: Sparse Masked Brain Modeling with Double-Conditioned Latent Diffusion Model for Human Vision Decoding, a framework that exploits the power of large-scale representation learning and mimics the sparse coding of information in the brain~\cite{sparse_cortex}, including the visual cortex~\cite{vinje2000sparse}.
Different from~\cite{maeHe}, we use a much larger representation-to-data-space ratio to boost the information capacity of learned representations.
Our contributions are as follows:
\begin{itemize}
\setlength\itemsep{0.05em}
\item We propose Sparse-Coded Masked Brain Modeling (SC-MBM), designed under biological guidance as an effective brain feature learner for vision decoding. 
\item Augmenting the latent diffusion model with double conditioning (DC-LDM), we enforce stronger decoding consistency while allowing variance under the same semantics.  
\item Integrating the representation ability of SC-MBM with the generation ability of DC-LDM, \textbf{\methodname} generates more plausible images with better preserved semantic information compared with previous methods. 
\item Quantitative and qualitative tests are performed on multiple datasets, including a new dataset that has not previously been used to evaluate this task. 
\end{itemize}

\section{Related Work}
\label{sec:related}

\paragraph{Conventional Decoding Methods} 
Conventional methods rely on training with fMRI and corresponding hierarchical image features extracted by a pre-trained VGG~\cite{kam2017,shen2019}. 
During testing, the predicted image features will either be used for classification or fed into a generative model like GAN~\cite{shen2019gan} to reconstruct the original stimulus.
Instead of directly learning the limited training pairs, \cite{guy2019} enabled unsupervised learning on unpaired fMRI and images with a reconfigurable autoencoder design. \cite{guy2022} further extended this method to images from diverse semantic categories. However, just as with conventional approaches, fMRI is used directly for training and decoding. 
In \cite{biggan2020,icgan2022}, a regression model was used to extract latent fMRI representation, which was then used to finetune a pre-trained conditional bigGAN for image decoding.  Mind Reader~\cite{lin2022mind} encoded fMRI signals into a pre-aligned vision-language latent space and used StyleGAN2 for image generation. These methods generate more plausible and semantically meaningful images. 
We note that there is parallel work to ours by Takagi and Nishimoto~\cite{fang2020reconstructing}, who proposed a method for image reconstruction from fMRI using Stable Diffusion. Their approach involves decoding brain activities to text descriptions and converting them to natural images using stable diffusion.


\paragraph{Masked Signal Modeling} 
The power of MSM in learning representations from a large-scale dataset was first exploited in \cite{bert}, which was later adapted to computer vision~\cite{maeHe, simmim, hog}. 
Successful applications to downstream tasks show that useful context knowledge is learned with MSM as a pretext task. 
In essence, MSM is a generalized denoising autoencoder that aims to recover the original data from the remaining after masking~\cite{cao2022understand}. 
The portion of data to mask is different across data modalities, with an extremely high mask ratio ($75\%$) usually used for visual signals~\cite{maeHe}. In contrast, due to the disparity in information density, a low mask ratio ($25\%$) is used in natural languages~\cite{bert}. 


\begin{figure*}[ht]
\setlength{\abovecaptionskip}{-1pt}
\setlength{\belowcaptionskip}{-15pt}
  \includegraphics[width=\textwidth]{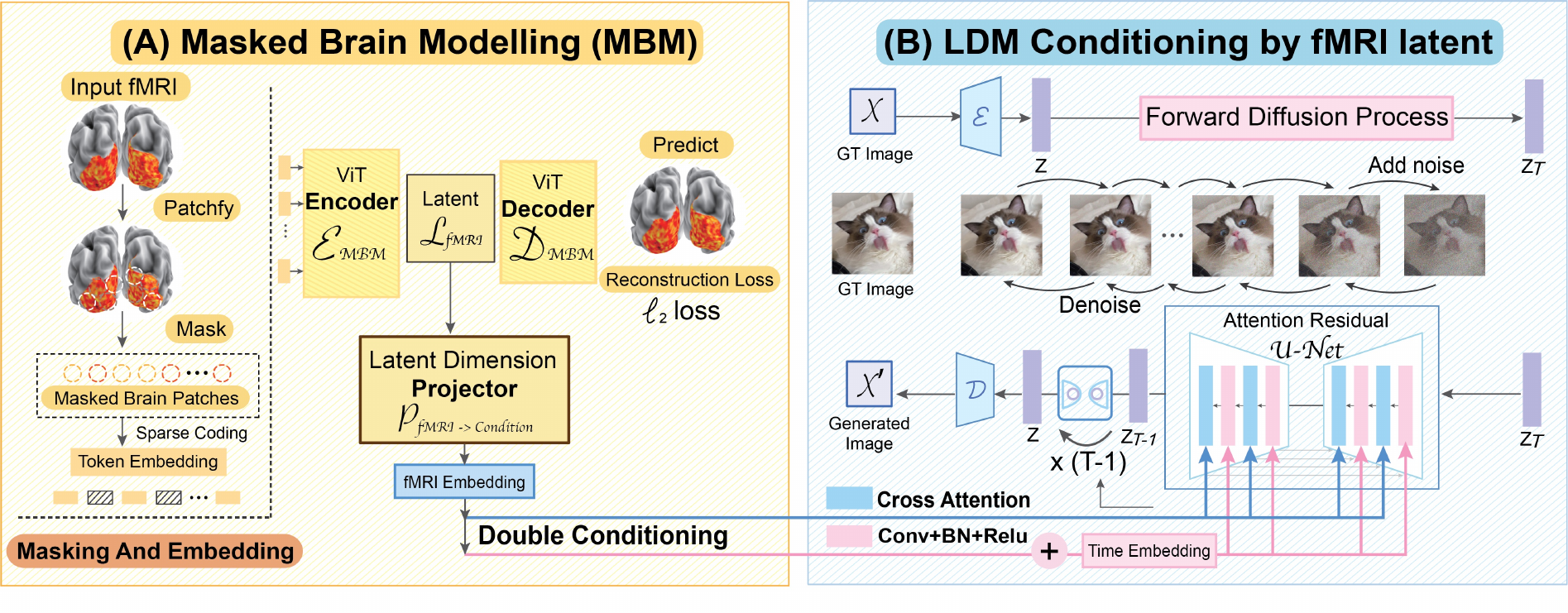}
  \centering
  \caption{\textbf{MinD-Vis}. 
\textbf{Stage A (left):} Pre-train on fMRI with SC-MBM. 
We patchify, randomly mask the fMRI, and then tokenize them to large embeddings. 
We train an autoencoder \footnotesize{($\mathcal{E_{MBM}}$ and $\mathcal{D_{MBM}}$)} \small to recover the masked patches.
\textbf{Stage B (right):} Integration with the LDM through double conditioning. 
We project the fMRI latent \footnotesize{($\mathcal{L}_{fMRI}$)} \small through two paths to the LDM conditioning space with a latent dimension projector \footnotesize{($\mathcal{P}_{fMRI\rightarrow Cond}$)}. \small 
One path connects directly to cross-attention heads in the LDM. 
Another path adds the fMRI latent to time embeddings. 
The LDM operates on a low-dimensional, compressed version of the original image (\ie image latent), however, the original image is used in this figure for illustrations.}
     \label{fig:flowchart}
\end{figure*}

\paragraph{Diffusion Probabilistic Models} 
Diffusion models~\cite{diffusion_models} are emerging generative models that generate high-quality content.
In its basic form~\cite{DDPM}, the diffusion model is a probabilistic model defined by a bi-directional Markov Chain of states.  
Two processes are transiting through the chain: 
\emph{(i)} The forward diffusion process gradually adds noise to the data until it is fully destroyed to an isotropic Gaussian noise; 
\emph{(ii)} The reverse process recovers the corrupted data by modeling a posterior distribution $p(x)$ at each state and eventually obtains a sample in the original data distribution~\cite{diffusion_models,DDPM,diffsong}. 
Formally, assume a Markov Chain with a fixed length $T$, then the reverse conditional probability can be expressed as $q(x_{t-1}|x_t)$, where $t=1,...,T$ and $x_t$ is obtained by corrupting the image $x_{t-1}$ with Gaussian noise. 
After parameterization, this conditional probability can be learned by optimizing a variational lower bound which can be simplified to the following objective~\cite{DDPM}:
\begin{equation}
  L_t^{simple}=\mathbb{E}_{x,\epsilon\sim\mathcal{N}(0,1),t}\left[\parallel\epsilon-\epsilon_\theta(x_t,t)\parallel_2^2\right],
  \label{eq:diff_obj}
\end{equation}
where $\epsilon_\theta(x_t,t)$ is a set of denoising functions that are usually implemented as UNets~\cite{unet,openaiDiff,ldm2022}. 
We refer readers to~\cite{DDPM} for detailed descriptions of the diffusion models.

\paragraph{Latent Diffusion Model (LDM)} 
Apart from the conventional diffusion models that generate samples in the original data space, another category of diffusion models that generate samples in the latent feature space has been proposed~\cite{d2c,ldm2022}. 
Operating in the latent feature space reduces the computational cost and introduces less spatial downsampling, giving better image synthesis quality.
The LDM proposed in \cite{ldm2022} consists of two components: (i) Vector Quantization (VQ) regularized~\cite{tamming} autoencoder that compresses images into lower-dimensional latent features and then reconstructs the images from features in the same space; 
(ii) UNet-based denoising model with attention modules.
Incorporating attention mechanisms into the UNet allows the flexibility to condition image generation through key/value/query vectors during the Markov Chain transitions. 

\section{Methodology}
\label{sec:method}

\subsection{Motivation and Overview}
In this subsection, we provide a detailed analysis of the fMRI data and elaborate on the motivations of our designs. 

\textbf{(i)} fMRI measures the brain blood-oxygen-level-dependent (BOLD) changes as 3D voxels that serve as a proxy for the underlying changes in brain activity.
Neighboring voxels often have similar amplitudes, indicating spatial redundancy in fMRI~\cite{fmri_redundancy}.

\textbf{(ii)} fMRI data is averaged across the time during which the stimulus is presented. A region of interest (ROI) of the averaged data is usually extracted as a \textbf{1D vector} of voxels (in the visual processing hierarchy). 
The ROI size (voxel number) is generally smaller than the image size (pixel number).
For example, \cite{kam2017} has about 4500 voxels (visual cortex), which is much smaller than a $256\times256$ RGB image. 
This creates a large difference in dimensionality when transforming fMRI into images.    

\textbf{(iii)} fMRI data from different datasets may have significant domain shifts due to experimental conditions and scanner setups. 
Even with the same scan conditions, ROI size and location mismatch persist due to individual differences (See~\cref{fig:individual_diff}). 


Driven by this analysis, we propose \textbf{\methodname}, designed with two sequential stages as outlined in \cref{fig:flowchart}. 
Briefly, in \textbf{Stage~A}, fMRI representations are learned by an autoencoder trained in a large fMRI dataset with masked signal modeling as a pretext task. The learned representations will be used as a condition to guide the image-generation process in the next stage. 
In \textbf{Stage~B}, the pre-trained fMRI encoder is integrated with the LDM through cross-attention and time-step conditioning for conditional synthesis. In this stage, the encoder is jointly finetuned with cross-attention heads in the LDM using paired annotations.

\subsection{Stage~A: Sparse-Coded MBM (SC-MBM)} 
Activity in the human brain involves non-linear interactions among 86 billion neuronal cells in the brain and are thus highly complex~\cite{olshausen1996emergence, rolls1995sparseness}. 
The fMRI measuring the BOLD signals is an indirect and aggregate measure of neuronal activities, which can be analyzed hierarchically with functional networks~\cite{fmri_hier_na, fmri_hier2, fmri_hier3}. 
These functional networks comprised of voxels of fMRI data have implicit correlations with each other in response to external stimuli~\cite{func_fmri, func_fmri_zhou}.
Therefore, learning these implicit correlations by recovering masked voxels will equip the pre-trained model with a deep contextual understanding of the fMRI data. 

\begin{figure}
\setlength{\belowcaptionskip}{-20pt}
    \centering
    \includegraphics[width=\linewidth]{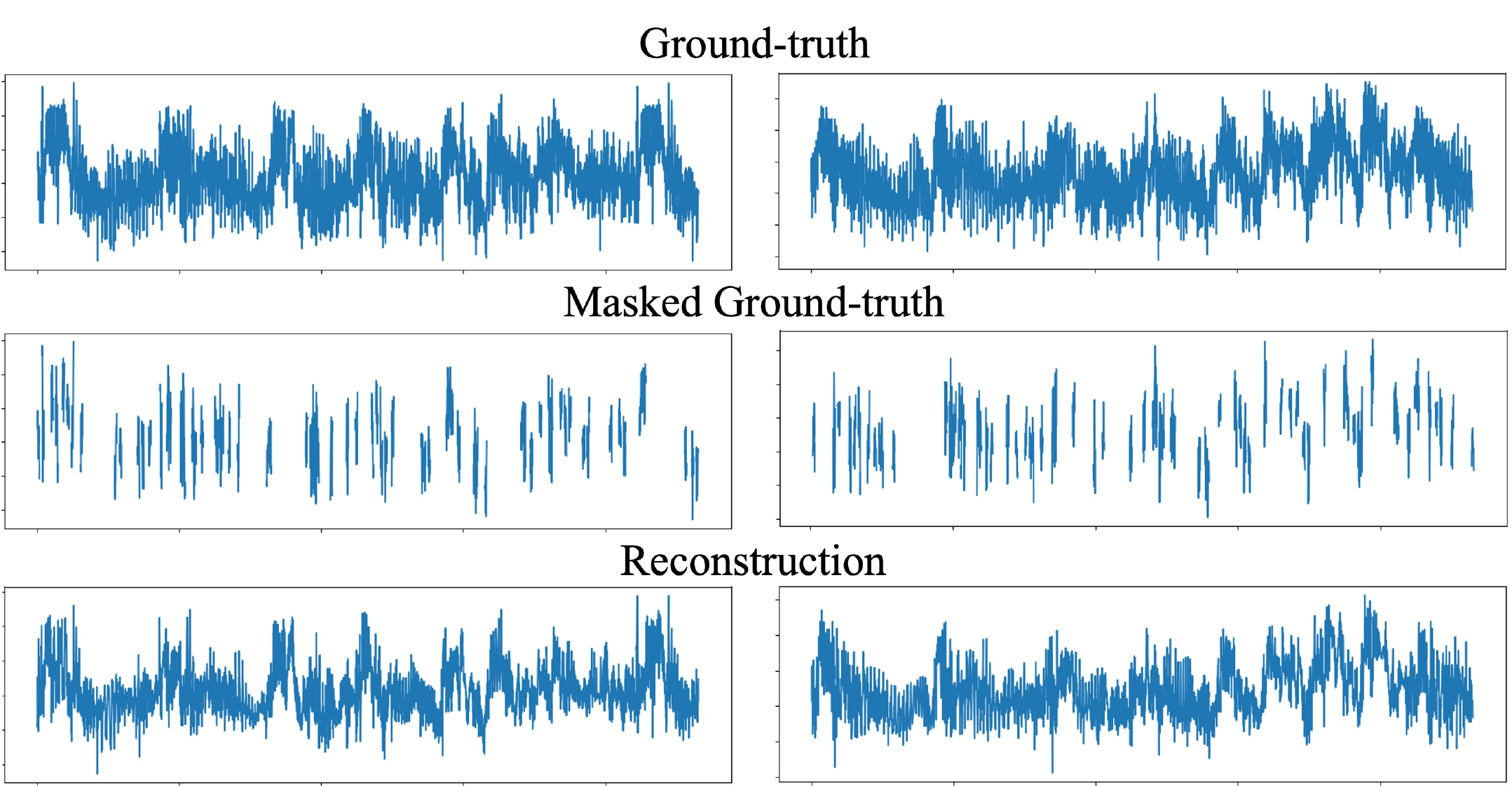}
    \caption{\textbf{Masked Brain Modeling.} Mask ratio 0.75; 4500 voxels}
    \label{fig:mbm_example}
\end{figure}

Following \cite{maeHe}, we divide the vectorized voxels into patches which will be subsequently transformed into embeddings using a 1D convolutional layer with a stride equal to the patch size.
The hemodynamic response and spatial smoothing functions in fMRI BOLD signal jointly cause spatial blurring, which creates spatial redundancy in fMRI data, like in natural images~\cite{engel1997retinotopic, shmuel2007spatio}.
Due to the spatial redundancy, fMRI data can still be recovered even if a large portion is masked (See \cref{fig:mbm_example}).
Thus, in the first stage of MinD-Vis, we can mask a large portion of the fMRI patches to save computations without losing the learning power of masked modeling.

Masked Image Modeling (MIM) uses the embedding-to-patch-size ratio around one~\cite{maeHe}, leading to a representation size similar to the original data size.
However, we use a large embedding-to-patch-size ratio, which significantly \textbf{increases the information capacity} with a large fMRI representation space. 
This design also relates to the sparse coding of information in the brain, which has been proposed as a general strategy for the representation of sensory information\cite{sparse}. 

We also adopt an asymmetric architecture as in\cite{maeHe}: the encoder is optimized to learn effective fMRI representations, while the decoder tries to predict the masked patches. Therefore, we make the decoder small in size, and it is discarded in Stage~B as long as the pre-training converges.

\paragraph{Visual Encoding and Brain-Inspired Sparse Coding} 

Here, we explain the biological basis of using SC-MBM to learn representations of visual stimuli in the brain from the perspective of visual encoding mechanisms.
Theoretical and empirical studies suggest that visual stimuli are sparsely encoded in the primary visual cortex~\cite{olshausen1996emergence, vinje2000sparse,rao1999predictive}, with most natural images activating only a portion of the neurons in the visual cortex. 
This strategy increases information transmission efficiency and creates minimal redundancy in the brain~\cite{rao1999predictive}.
As a result, visual information of natural scenes can be reconstructed from a small portion of data collected from the primary visual cortex via different imaging modalities, including fMRI~\cite{yoshida2020natural, freeman2013functional}. 
This observation is interesting for the computer vision community because the sparse coding could be an efficient way for vision encoding in computer vision as well~\cite{sparse,sparseCode}. 

Sparse coding is an encoding strategy that in essence uses over-complete bases to represent data, where more locality is generally enforced to generate smoother representations~\cite{sparselocality,sparselocality2}. 
In SC-MBM, fMRI data are divided into patches to introduce locality constraints. Then each patch is encoded into a high-dimensional vector space with a size much larger than the original data space, thus creating an over-complete space for fMRI representation (See Appendix). 
Emulating the brain vision encoding, SC-MBM can be a biologically-valid and effective brain feature learner for fMRI decoding.

\begin{figure*}[t]
\setlength{\belowcaptionskip}{-15pt}
  \centering
   \includegraphics[width=\linewidth]{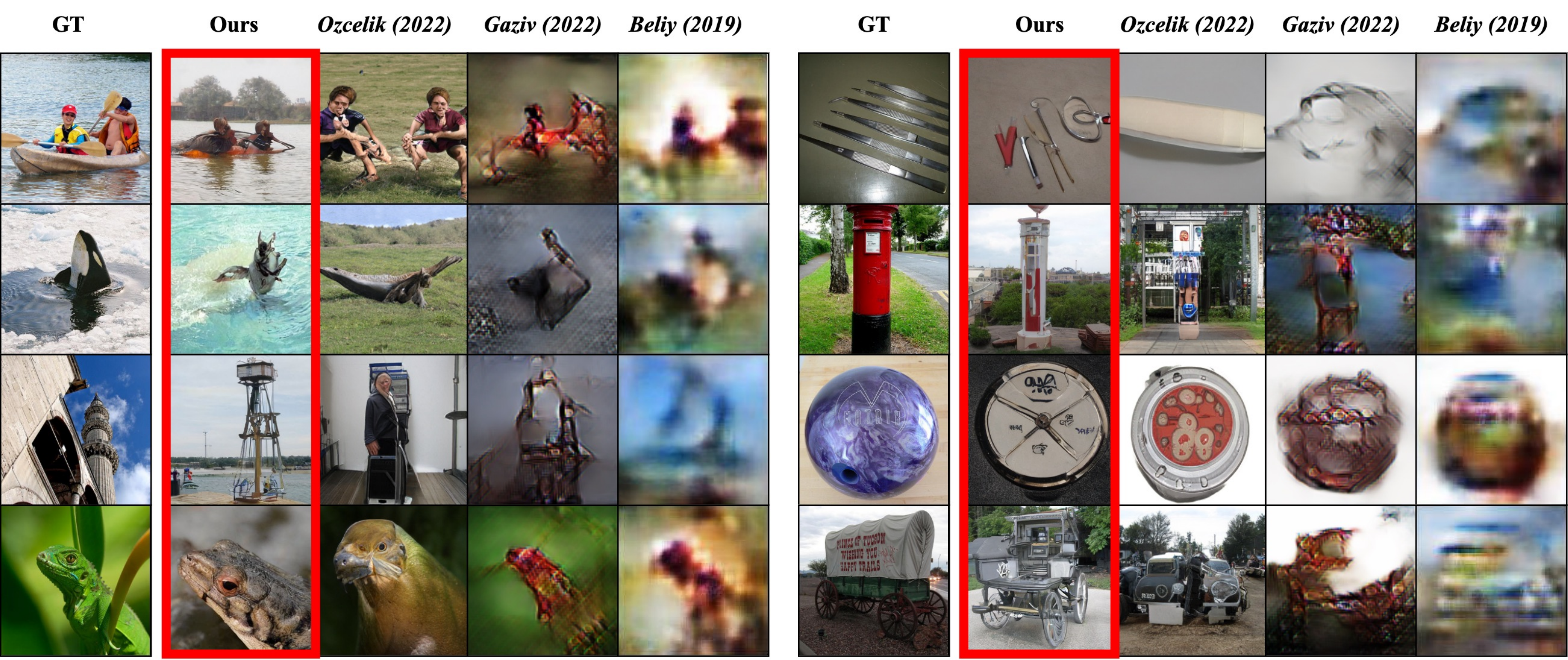}
   \caption{\textbf{Decoding Performance Comparisons on GOD Test Set.} 
   The ground truth, images reconstructed by \methodname\, and images reconstructed from three other methods are shown for comparison.
  \methodname\ decoded the most accurate and plausible images with semantically similar details.}
   \label{fig:compare_figs}
\end{figure*}

\begin{figure}
\setlength{\abovecaptionskip}{-1pt}
\setlength{\belowcaptionskip}{-15pt}
  \centering
    \includegraphics[width=\linewidth]{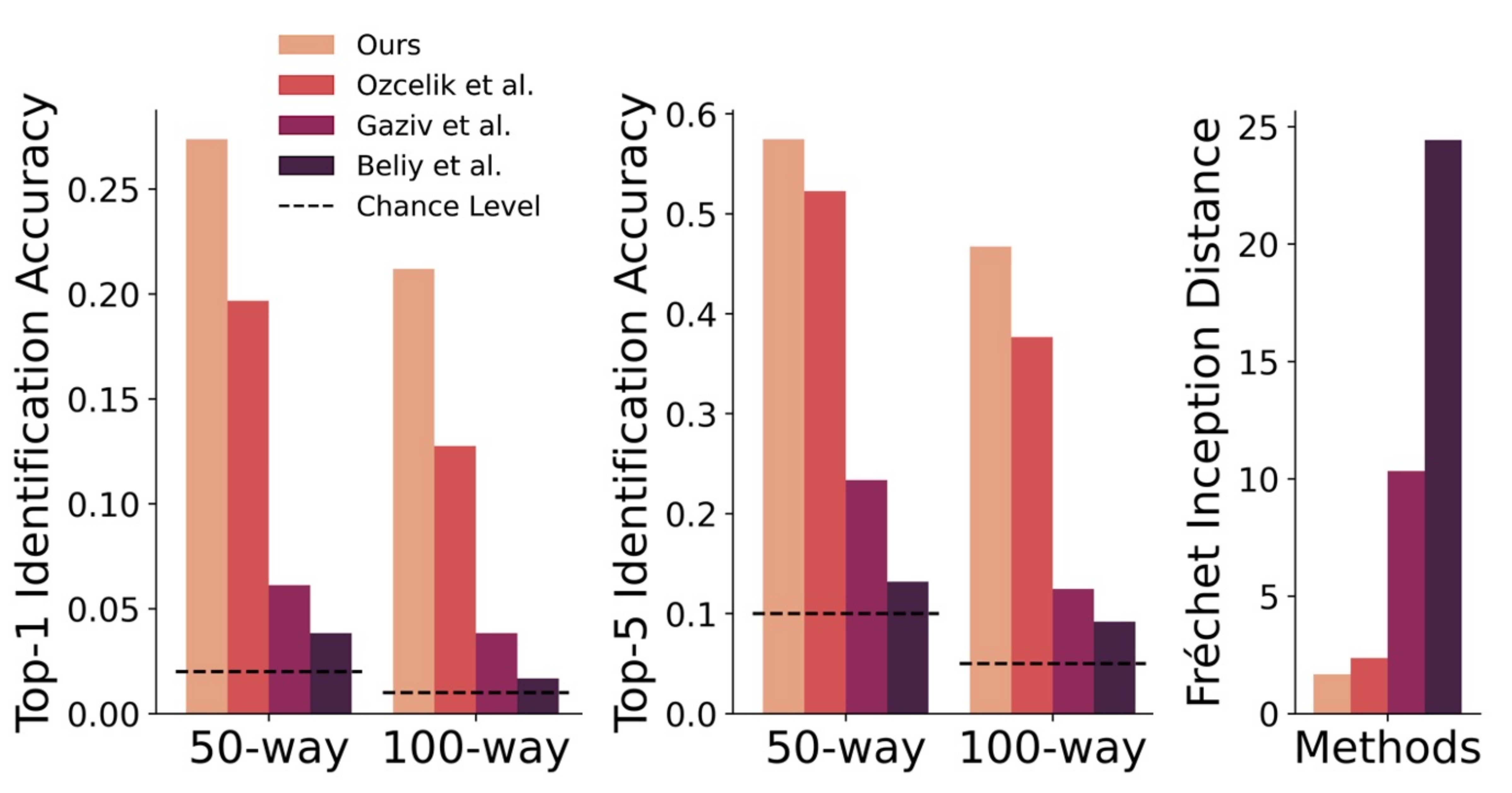}
  \caption{\textbf{Quantitative Performance Comparisons on GOD Test Set.}
  Performance is evaluated in terms of semantic correctness
 (\emph{1000-trial n-way top-k classification accuracy; the higher the better}) and generation quality (\emph{FID; the lower the better}). 
 }
  \label{fig:class_acc}
\end{figure}

\subsection{Stage B: Double-Conditioned LDM (DC-LDM)} 
After the large-scale context learning in Stage~A, the fMRI encoder transforms fMRI data into sparsely coded representations with locality constraints. 
To further decode visual contents from this abstract representation and allow for sampling variance, 
we formulate the decoding task as a conditional synthesis problem and approach it with a pre-trained LDM. 

The LDM operates on the image latent space denoted by $\mathcal{E}(x)$ where $x$ is an image in pixel space and $\mathcal{E}(\cdot)$ is a VQ encoder. 
In our setting, we omit $\mathcal{E}(x)$ and use $x$ directly to represent the latent variable of LDM for simplicity. 
Specifically, given the fMRI data $z$, we aim to learn the reverse diffusion process formulated by $q(x_{t-1}|x_t, z)$. 
As proposed in \cite{ldm2022}, conditional information is applied through cross-attention heads in the attention-based UNet, where \mbox{CrossAttention}$(Q,K,V)=$ \mbox{softmax}$\left(\frac{QK^T}{\sqrt{d}}\right)$, with 
\begin{equation*}
    Q=W^{(i)}_Q\varphi_i(x_t),~K=W^{(i)}_K\tau_\theta(z),~V=W^{(i)}_V\tau_\theta(z).
\end{equation*}
Here, $\tau_\theta$ is the fMRI encoder with a suitable dimension projector, $\varphi_i(x_t)$ denotes intermediate values of the UNet and $W^{(i)}_Q$, $W^{(i)}_K$, $W^{(i)}_V$ are projector matrices with learnable parameters. 

Diversity and consistency are two opposite objectives when sampling a conditional generative model.
Sampling diversity across various modalities such as label-to-image and text-to-image is very important in many image-generation tasks.
However, the fMRI-to-image transition relies more on \textbf{generation consistency}---decoded images from similar brain activities are expected to be semantically similar. 
Thus, a stronger conditioning mechanism is desired to ensure such generation consistency, especially for probabilistic diffusion models. 

In this way, we integrate the cross-attention conditioning with another conditioning method called the \emph{time steps conditioning}~\cite{openaiDiff} to provide stronger guidance for our task. 
In time steps conditioning, we add $\sigma_\theta(\tau_\theta(z))$ to time step embeddings, where $\sigma_\theta(\cdot)$ is another suitable dimension projector. 
Time step embeddings are used in intermediate layers of the UNet, thus we have $\varphi_i(x_t) = \varphi_i(x_t, \sigma_\theta(\tau_\theta(z)))$. 
We further reformulate the optimization objective \cref{eq:diff_obj} to a \emph{double conditioning} alternation: 
\begin{equation}
    \label{eq:diff_obj_con}
    L_t^{cond}=\mathbb{E}_{x,\epsilon\sim\mathcal{N}(0,1),t}\left[\parallel\epsilon-\epsilon_\theta(x_t,t,\tau(z),\sigma(\tau(z)))\parallel_2^2\right].
\end{equation}
We omit the parameterization symbol $\theta$ in $\tau(\cdot)$ and $\sigma(\cdot)$ for simplicity. 
Additionally, we have $\tau(z)\in\mathbb{R}^{M\times d_\tau}$ and $\sigma(\tau(z))\in\mathbb{R}^{1\times d_t}$, where $d_\tau$ and $d_t$ are the latent dimensions and time embedding dimension respectively, and $M$ is a tunable parameter.

\paragraph{Finetuning} 
After the fMRI encoder is pre-trained with SC-MBM, it is integrated with a pre-trained LDM through double conditioning. 
Commonly, the encoder's output is averaged, or a \emph{cls} token is appended to produce a pooled 1D feature vector for downstream tasks~\cite{bert,maeHe}. This strategy is effective for tasks like prediction and classification, where learned knowledge is expected to be distilled, producing distinguishable features. 
However, pooling into a 1D vector is inappropriate for retaining fMRI representations' sparsity and information capacity. 
Instead, we used convolution layers to pool the encoder's output into a latent dimension of $\mathbb{R}^{M\times d_\tau}$ as described in \cref{eq:diff_obj_con}.

The fMRI encoder, cross-attention heads, and projection heads are jointly optimized, while other parts are fixed. 
Finetuning the cross-attention heads is critical for bridging the pre-trained conditioning space and fMRI latent space. 
The finetuning is performed end-to-end with fMRI-image pairs, during which a clearer connection between the fMRI and image features will be learned through the large-capacity fMRI representations.  


\section{Experiments}
\label{sec:experiment}

\subsection{Datasets and Implementation} 
\paragraph{Datasets}
Three public datasets were used in this study: 
Human Connectome Project (HCP) 1200 Subject Release~\cite{hcp}; 
Generic Object Decoding Dataset (GOD)~\cite{kam2017};
and Brain, Object, Landscape Dataset (BOLD5000)~\cite{chang2019bold5000}. 
Our upstream pre-training dataset comprised fMRI data from HCP and GOD.
Combining these two, we obtained 136,000 fMRI segments from 340 hours of fMRI scan, which is, by far, the largest fMRI pre-training dataset in the fMRI-image decoding task. 
The HCP dataset is commonly used in neuroscience research, containing only fMRI data. 
While the GOD is an fMRI-image paired dataset designed for fMRI-based decoding. 
The pairs in GOD were used for finetuning in our main analysis. The GOD consists of 1250 different images from 200 distinct classes, in which 1200 images were used as the training set, and the remaining 50 images were used as the testing set. The training set and testing set have no overlapping classes.  
The BOLD5000 dataset was used as the validation dataset in our study. It consists of 5254 fMRI-image pairs from 4916 distinct images, 113 images of which are used for testing. This is the first time that the BOLD5000 is used for fMRI decoding tasks.  

\paragraph{Implementation}
The fMRI pre-training model is similar to ViT-Large~\cite{vit} with a 1D patch embedder. 
We used a patch size of 16, embedding dimension of 1024, encoder depth of 24, and mask ratio of 0.75 as our Full model setting 
with an ImageNet class-conditioned pre-trained LDM. 
Different parameter choices are explored in our ablation study. 
Unless stated otherwise, the Full model is pre-trained for 500 epochs and finetuned for another 500. 
Results from the best model are reported. Images are generated at a resolution of $256\times256$ with 250 PLMS steps~\cite{plms}.
See Appendix for dataset and implementation details.

\subsection{Evaluation Metric} 
\paragraph{N-way Classification Accuracy}
Following \cite{guy2022}, we used the $n$-way top-1 and top-5 accuracy classification task to evaluate the semantic correctness of our results, where for multiple trials, top-1 and top-5 classification accuracies were calculated in $n-1$ randomly selected classes plus the correct one. 
\emph{Note that we did not consider the pixel-level metrics as we aimed to recover the semantically correct images in this work.} 

In \cite{guy2022}, the authors generated a typical feature for each class selected and compared the distance between the reconstructed images and the typical features. 
However, this metric in \cite{guy2022} is hard to reproduce, and the semantic classification result largely depends on how the features are computed. 
Therefore, we propose a more straightforward and reproducible method, where a pre-trained ImageNet1K classifier~\cite{vit,resnetwight} is used
to determine the semantic correctness of generated images rather than handcrafted features. We describe this evaluation method in \cref{alg:metric}. 
Specifically, both ground-truth and generated images are input to the classifier first. Then we check for the generated image if the top-$k$ classification in $n$ selected classes matches the ground-truth classification. 
This metric does not require the ground-truth image to be from the ImageNet 1k classes. As long as semantic classification results of the ground-truth and the generated image match, it will be considered to be correct. 

\paragraph{Fréchet inception distance (FID)} 
The FID~\cite{heusel2017gans} is a commonly used metric to assess image generation quality. 
In our experiments, we measured the FID between ground-truth images and generated images in the testing set. 
Note that FID is only used as a reference in our experiments due to the limited number of images available in GOD, which may lead to an underestimated distribution.

\section{Results}
\label{sec:results}

Our main results are based on GOD which has no overlapping classes in the training and testing set. The training and testing were performed on the same subject, as individual differences remain a barrier when decoding at the group level~\cite{kam2017,guy2019,guy2022,biggan2020,icgan2022}.
To compare with the literature, we report results from Subject~3 here and leave other subjects in the Appendix.

We compared our results with Ozcelik~\etal~\cite{icgan2022}, Gaziv~\etal~\cite{guy2022} and Beliy~\etal~\cite{guy2022}. 
Gaziv~\etal and Beliy~\etal used the conventional method, which decoded images with higher pixel similarity but less plausibility and semantic details.
On the other hand, Ozcelik~\etal generated more plausible and semantically meaningful images using a pre-trained GAN.
Based on the best-reconstructed samples of these methods (resized to $256\times256$), we performed a 1000-trial, $n$-way top-$k$ accuracy identification task as described in \cref{alg:metric}. The experiment is repeated for $n=50, 100$ and $k=1, 5$ in the GOD testing set.

From \cref{fig:class_acc}, our identification accuracy outperformed the Ozcelik \etal in the 50-way top-1 accuracy task by $39\%$ and in the 100-way top-1 accuracy task by $66\%$, achieving a success rate of 0.274 and 0.212 respectively. 
The generated images from Gaziv~\etal and Beliy~\etal were close to the ground-truth at the pixel level but contained few semantically meaningful details, as could be observed in \cref{fig:compare_figs}. 
For example, our method generated plausible details such as water and waves in the first and second images, drawings on the bowling ball, wheels of the carriage, \etc, which were not present in the previous decoded images. 
The image quality is also reflected by the FID, where we achieved 1.67 with our best samples, while Ozcelik~\etal and others achieved 2.36 or more with the best samples generated by their method.  
Interestingly, color mismatches are observed in some cases with the color difference well preserved. 
It can be explained with~\cite{bird2014categorical} which suggests the color category information is processed in the frontal lobes as a cognitive process, while the visual cortex only recognizes the difference in colors.

\begin{figure}[t]
\setlength{\belowcaptionskip}{-15pt}
  \centering
   \includegraphics[width=\linewidth]{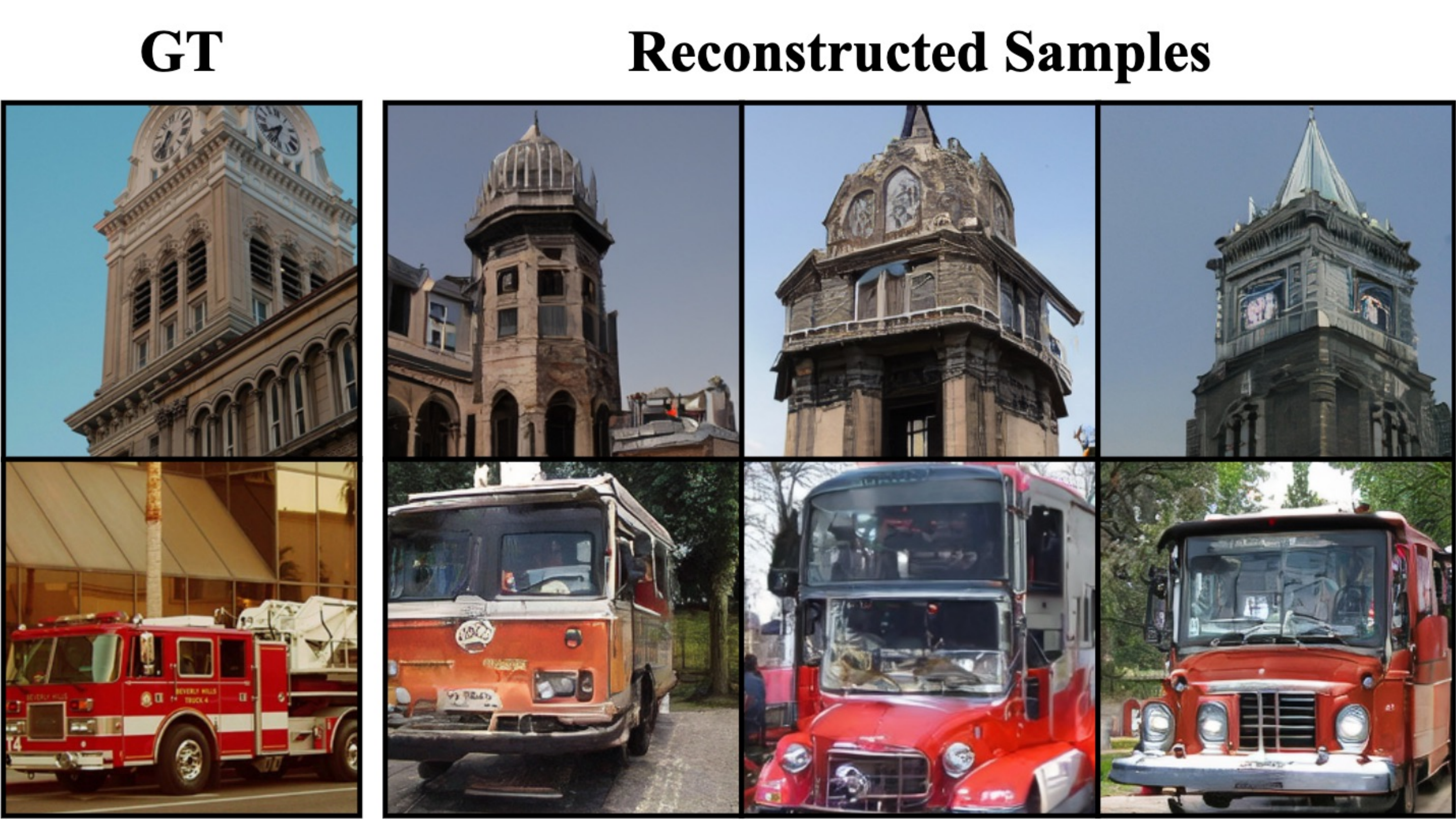}
   \caption{\textbf{Generation Consistency of MinD-Vis}. Images generated by our method were consistent across different samplings trials, sharing similar low-level features and semantics.}
   \label{fig:consistent}
\end{figure}

\subsection{Generation Consistency}
The consistency of our method was tested by decoding the same fMRI data multiple times with different random states.
Five samplings with different random states were performed in the testing set for each fMRI. 
In the 50-way and the 100-way top-1 accuracy identification tasks, we achieved an average success rate across the five samplings of $0.2385 \scriptstyle{\pm 0.030}$ and $0.1736 \scriptstyle{\pm 0.029}$ respectively, which are statistically higher than the best sampling results from Ozcelik \etal by $21\%$ and $35\%$. Regarding image quality, we achieved an average FID of $2.22 \scriptstyle{\pm 0.3}$ across the five samplings.  
The standard deviations across 5 samplings indicate that the generated images will always be in the same semantic category.
It can also be seen in \cref{fig:consistent} where isomorphic samplings share similar details such as shape, color, texture, and semantics, matching with the ground-truth across trials.

\subsection{SC-MBM Design}
This section will discuss the ablation study on the SC-MBM pre-training stage with various important parameters.
Results are summarized in \cref{tab:MBM_test}. 
For all experiments in this section, the 50-way, top-1 accuracy semantic identification task was performed with the best models obtained from the finetuning of 500 epochs. 
Average results over five samplings were reported.

\paragraph{Testing Without SC-MBM} 
To show that useful representations were learned with SC-MBM, we trained two models directly using the fMRI-image pairs without the SC-MBM pre-training. 
The first model consisted of an untrained fMRI encoder with the same architecture as the Full model. 
The second model consisted of an untrained fMRI encoder with a depth of only 2. 
The second model was designed to have fewer parameters, making it less likely to overfit the data. 
All the other settings were the same. 
The results correspond to Model 1 and 2 in \cref{tab:MBM_test}, where the Full model significantly outperformed the other two models without the SC-MBM pre-training, showing that the pre-training is crucial. 
In fact, without SC-MBM these two models even failed to generate sensible images (See Appendix).

\begin{table}[ht]
\setlength{\belowcaptionskip}{-15pt}
\resizebox{\columnwidth}{!}{%
\begin{tabular}{ccccc}
\toprule
Model & \begin{tabular}[c]{@{}c@{}}Embedding\\ Dim\end{tabular}& \begin{tabular}[c]{@{}c@{}}Mask\\ Ratio\end{tabular} & Params & Acc (\%) \\ \midrule
\textbf{Full} & \textbf{1024} & \textbf{0.75} & \textbf{303M} & \textbf{23.9\scriptsize{$\pm3.00$}} \\ \midrule\midrule
1 & \multicolumn{2}{c}{w/o SC-MBM + same Encoder} & \textbf{303M} & \cellcolor[HTML]{CBCEFB}2.6   \scriptsize{$\pm1.39$}\\
2 & \multicolumn{2}{c}{w/o SC-MBM + smaller Encoder} & \textbf{25M} & \cellcolor[HTML]{CBCEFB}3.4 \scriptsize{$\pm0.86$}\\ \midrule
3 & \textbf{32} & 0.75 & 0.3M & \cellcolor[HTML]{CBCEFB}5.4  \scriptsize{$\pm1.50$}\\
4 & \textbf{64} & 0.75 & 1.2M & \cellcolor[HTML]{CBCEFB}6.9  \scriptsize{$\pm1.10$}\\
5 & \textbf{128} & 0.75 & 4.7M & \cellcolor[HTML]{FFCCC9}14.8  \scriptsize{$\pm1.78$}\\
6 & \textbf{256} & 0.75 & 18.9M & \cellcolor[HTML]{FFCCC9}15.9  \scriptsize{$\pm1.70$}\\
7 & \textbf{512} & 0.75 & 75.6M & \cellcolor[HTML]{FFCCC9}17.9  \scriptsize{$\pm2.58$}\\
8 & \textbf{768} & 0.75 & 170M & \cellcolor[HTML]{FFCCC9}17.7  \scriptsize{$\pm1.42$}\\
9 & \textbf{1280} & 0.75 & 472M & \cellcolor[HTML]{FFCCC9}15.5 \scriptsize{$\pm3.83$}\\ \midrule
10 & 1024 & \textbf{0.35} & 303M & \cellcolor[HTML]{ACE1AF}19.6 \scriptsize{$\pm3.40$}\\
11 & 1024 & \textbf{0.45} & 303M & \cellcolor[HTML]{FFFC9E}20.0 \scriptsize{$\pm1.89$}\\
12 & 1024 & \textbf{0.55} & 303M & \cellcolor[HTML]{FFFC9E}18.1 \scriptsize{$\pm2.87$}\\
13 & 1024 & \textbf{0.65} & 303M & \cellcolor[HTML]{ACE1AF}21.7 \scriptsize{$\pm3.61$}\\
14 & 1024 & \textbf{0.85} & 303M & \cellcolor[HTML]{FFCCC9}16.1 \scriptsize{$\pm1.00$}\\ \bottomrule
\end{tabular}%
}
 \begin{tablenotes}
    \definecolor{purple}{HTML}{BFB9FA}  
    \definecolor{pink}{HTML}{F2ACB9}  
    \definecolor{yellow}{HTML}{FFBF00}  
    \definecolor{green}{HTML}{ACE1AF} 
    \scriptsize
        \item $^{\dagger}$ 
        $p<0.0001$ (\textbf{\textcolor{purple}{purple}});  
        $p<0.01$ (\textbf{\textcolor{pink}{pink}}); 
        $p<0.05$ (\textbf{\textcolor{yellow}{yellow}}); 
        $p>0.05$ (\textbf{\textcolor{green}{green}})
    \end{tablenotes}
\caption{\textbf{SC-MBM Ablation Results}. Params: trainable parameters in the fMRI encoder;
Cell colors reflect statistical significance differences (two-sample t-test) in accuracy compared with the Full model.}
  \label{tab:MBM_test}
\end{table}

\paragraph{Patch Embedding Dimension}
Boosting the fMRI representation size using a large patch embedding matches the sparse coding mechanism of underlying visual information processing in the brain. 
Moreover, using a large patch embedding increases the information capacity of the representation. 
But larger embedding means more training parameters leading to a more data-hungry model. 
To balance this tradeoff, we tested SC-MBM models with different patch embedding dimensions ranging from 32 to 1280 (Model 3-9 in \cref{tab:MBM_test}). 
We found that the accuracy generally increased as patch dimension increased, and accuracy peaked at $23.9\%$ with 1024 patch embedding dimensions (full model), after which accuracy decreased as patch dimensions increased further.

\paragraph{Mask Ratios}
We used a high mask ratio in SC-MBM due to high spatial redundancy in fMRI data. 
In \cref{tab:MBM_test} Model 10-14, we show that a high mask ratio does not impair the decoding performance initially, with the highest average accuracy achieved with a relatively high mask ratio of 0.75. 
Importantly, using a high mask ratio significantly reduces memory consumption since the encoder only operates over unmasked patches. 
This is an important consideration for fMRI as SC-MBM is more memory-intensive than MIM due to the higher embedding-to-patch-size ratio. 

\subsection{DC-LDM Finetuning Design}
This section will discuss the ablation study on the DC-LDM finetuning designs from three perspectives: conditioning methods, optimization designs, and pre-trained LDMs. 
Here, all ablations used the same pre-trained fMRI encoder as the Full model. 
Only important parameters in the finetuning stage were varied. 
The 1000-trial, 50-way, top-1 semantic identification test was performed. The results are summarized in \cref{tab:ldm_design}, where five different samplings were averaged for each condition. 

\begin{table}[ht]
\setlength{\belowcaptionskip}{-15pt}
\resizebox{\columnwidth}{!}{%
\begin{tabular}{ccccc}
\toprule
Model & Condition & Finetune & Pre-trained LDM & Acc (\%) \\ \midrule
\textbf{Full} & \textbf{C + T} & \textbf{E + A} & \textbf{Label2Image} & \textbf{23.9\scriptsize{$\pm3.00$}} \\ \midrule\midrule
1 & \textbf{C only} & E + A & Label2Image & \cellcolor[HTML]{FFCCC9}15.6 \scriptsize{$\pm0.69$}\\
2 & C+T & \textbf{E only} & Label2Image &   \cellcolor[HTML]{FFCCC9}13.76 \scriptsize{$\pm2.60$}\\
3 & C+T & E + A & \textbf{Text2Image} &     \cellcolor[HTML]{FFCCC9}13.42\scriptsize{$\pm3.00$}\\
4 & C+T & E + A & \textbf{Layout2Image} &   \cellcolor[HTML]{FFCCC9}15.99\scriptsize{$\pm3.00$}\\ \bottomrule
\end{tabular}%
}
 \begin{tablenotes}
    \definecolor{purple}{HTML}{BFB9FA}  
    \definecolor{pink}{HTML}{F2ACB9}  
    \definecolor{yellow}{HTML}{FFCC00}  
    \definecolor{green}{HTML}{ACE1AF} 
    \scriptsize
        \item $^{\dagger}$ 
        $p<0.0001$ (\textbf{\textcolor{purple}{purple}});  
        $p<0.01$ (\textbf{\textcolor{pink}{pink}}); 
        $p<0.05$ (\textbf{\textcolor{yellow}{yellow}); 
        $p>0.05$ (\textbf{\textcolor{green}{green}})}
\end{tablenotes}
 \caption{\textbf{DC-LDM Ablation Results}. 1: cross-attention condition only; 2: optimizing fMRI encoders only; 3: LDM pre-trained on text conditions (LAION); 4: LDM pre-trained on layout conditions (OpenImages). 
 Abbr.: C (Cross-attention condition); T (Time condition); E (Encoder); A (Cross-attention heads).
 Cell colors reflect statistical significance (two-sample t-test) in accuracy compared with the Full model.}
  \label{tab:ldm_design}
\end{table}

\paragraph{Conditioning Methods} Here, we showed that the double conditioning method increased the conditioning strength in \cref{tab:ldm_design}, where using only cross-attention conditioning achieved an identification accuracy of $15.6\%$ (Model~1), which was significantly lower than the full method. 

\paragraph{Optimizing LDM} 
We proposed to finetune the fMRI encoder and the cross-attention heads jointly because the LDM was pre-trained in a different conditioning space. 
For example, for the ImageNet class-conditioning pre-trained LDM, the cross-attention heads were pre-trained to receive the class label information. 
To justify this choice, we tested on a model with the fMRI encoder finetuned and the cross-attention heads untouched. 
As shown in Model 2 in \cref{tab:ldm_design}, the average identification accuracy dropped to $13.7\%$ when only the fMRI encoder was finetuned, indicating stronger semantic guidance with the double conditioning. 
The visual quality and correspondence to the ground-truth of the generated images also decreased significantly (See Appendix).  

\paragraph{Pre-trained LDM} 
The pre-trained LDM determines the model's generative ability and the conditioning latent space to which the fMRI encoder would adapt. 
We considered three pre-trained LDM provided by~\cite{ldm2022}, which were trained on datasets with different conditioning tasks, \ie ImageNet (label conditioning), LAION (text conditioning)~\cite{laion} and OpenImages (layout conditioning)~\cite{openimages}. 
As shown in Model 3-4 \cref{tab:ldm_design}, the ImageNet pre-trained LDM (used in the full model) showed the best performance in the same decoding task. 
Notably, images generated by models pre-trained on LAION and OpenImages were less visually favorable and plausible (See Appendix). 
This result is surprising because both LAION and OpenImages contain diverse images from various categories. 
We attribute the main reason for their poor performance to the complexity of their conditioning latent space. 
With limited training pairs, the class-conditioning latent space is easier to adapt to, compared with the latent space of the text-conditioning model and the layout-conditioning model.


\subsection{Replication Dataset}
\begin{figure}[t]
\setlength{\belowcaptionskip}{-15pt}
  \centering
   \includegraphics[width=\linewidth]{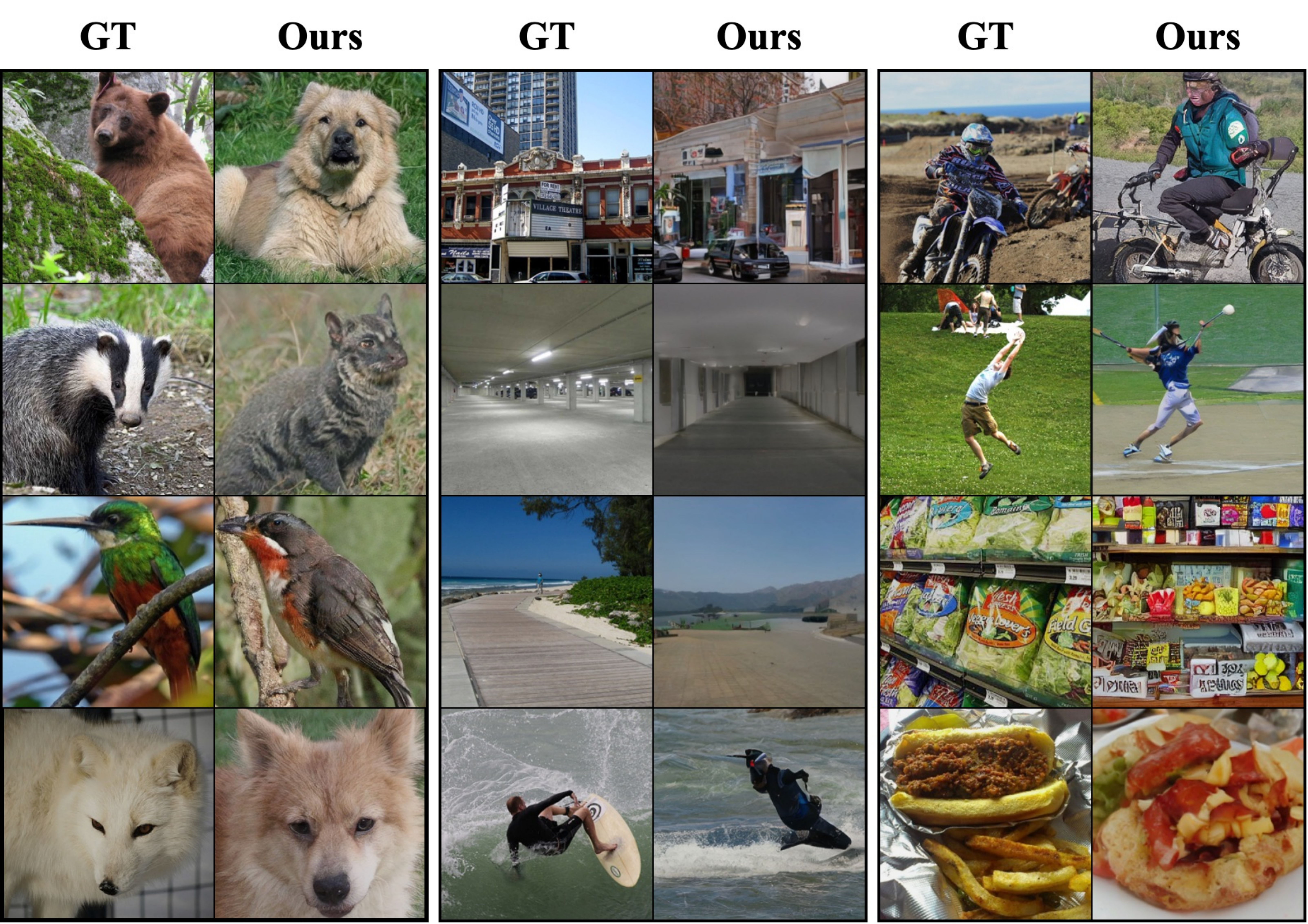}
   \caption{\textbf{Replication Dataset (BOLD5000)}. 
   It achieved similar quantitative results as the GOD dataset. 
   50-way top-1 identification accuracy: 34\%; FID: 1.2 (Subject~1).}
   \label{fig:bold5000}
\end{figure}

We validated our method on BOLD5000 using the same pre-trained fMRI encoder.
Similarly, the pre-trained encoder was firstly finetuned for 20 epochs in the testing set of BOLD500 with wrap-around paddings to compensate for the unequal ROI size from the pre-training set, after which the model is further tuned with the fMRI-image training pairs in BOLD5000.
All other settings were the same as the Full model.  
For the four subjects in BOLD5000, we achieved a $19\%$ to $34\%$ best accuracy in the 1000-trial, 50-way, top-1 accuracy semantic identification task (See Appendix).
The generated images matched the ground-truth stimulus in both semantics and low-level features (\cref{fig:bold5000}). 
Our model accurately reconstructs images containing objects and animals, architecture, and landscapes.  

\begin{figure}[t]
\setlength{\belowcaptionskip}{-15pt}
  \centering
   \includegraphics[width=\linewidth]{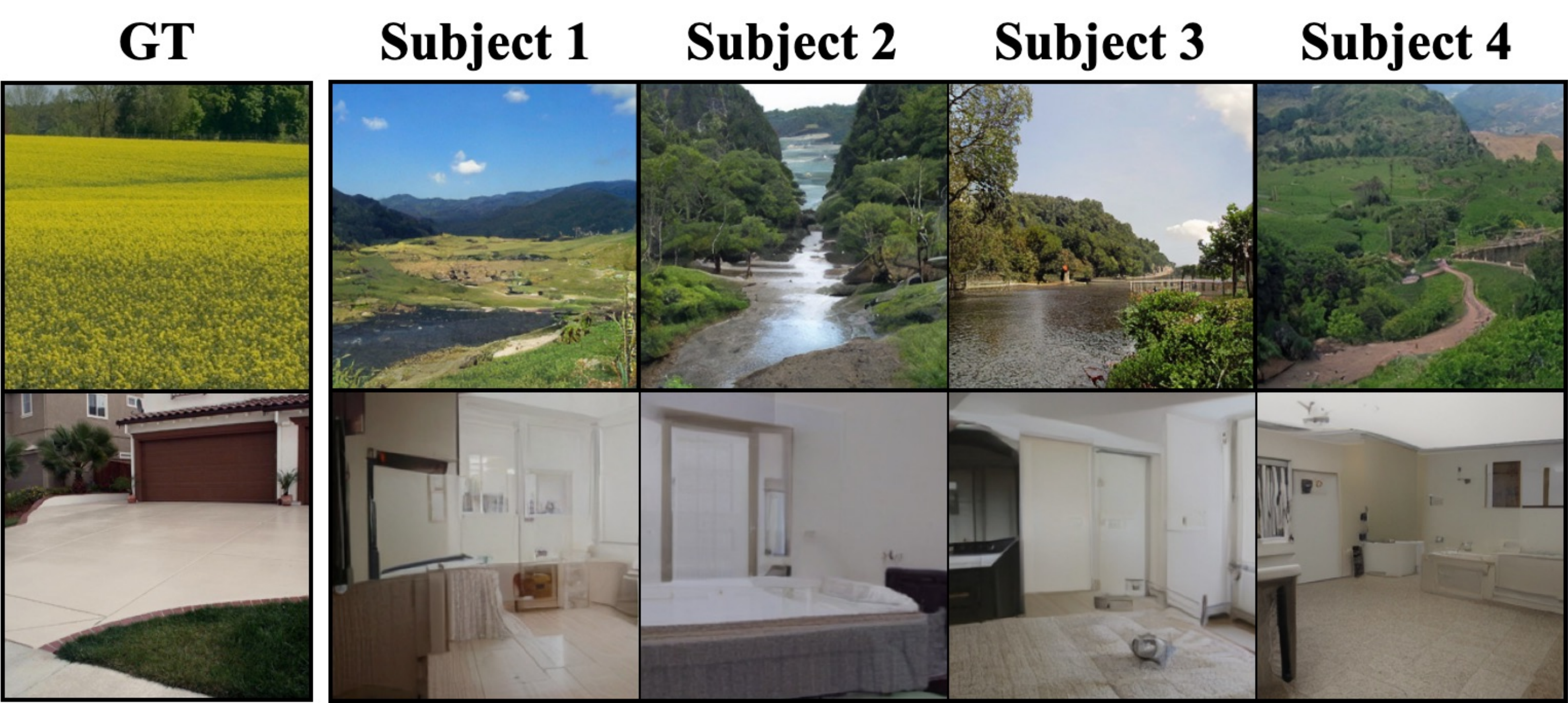}
   \caption{\textbf{Extra Features Decoded.} 
Imagery-related details can be decoded with our method.
\eg
the river and blue sky were decoded with natural scenery stimulus (top row); similar interior decorating of indoor environments was decoded when a house was presented (bottom row).
  }
   \label{fig:diff_subjects}
\end{figure}
Interestingly, we reconstructed similar images for some natural scenes with extra details that do not exist in the ground-truth stimulus. 
These extra details, for example, the river and the blue sky in \cref{fig:diff_subjects}, may reflect imagined scenery in the subject's mind when viewing the visual stimuli, which is captured in their brain activities.
As reported in \cite{shen2019,kam2017}, features of imaginary images can also be decoded from the visual cortex. 

To the best of our knowledge, this is the first work that performs fMRI decoding on BOLD5000. 
Additionally, adapting the same pre-trained model to this dataset shows that the SC-MBM pre-training indeed learns useful representations of brain recordings even when distinct domain shifts exist. 
These learned representations are shared and generalizable to datasets with different scanning protocols and preprocessing pipelines. 

\section{Discussion and Conclusion}
\label{sec:conclusion}






\paragraph{Limitations} 
{\methodname}, in its current form, lacks strong pixel-level guidance and interpretation analysis, which limits its pixel-level performance (see \ref{table:pixel-level-result}) and the biological understanding of the features learned by MBM.

\paragraph{Future Work}
Similar to all previous work, {\methodname} focuses on individual decoding using the visual cortex only. But as a complex cognitive process, human vision may be affected by regions beyond the visual cortex. Therefore, future studies should extend to cross-subject generalization and also the incorporation of other brain regions. 
Additionally, the two-stage decoupling design of {\methodname} allows us to explore the potential of emerging large-scale models and representation learning techniques in cognitive neuroscience, which is also subject to future studies.

\paragraph{Conclusion}
We proposed a two-stage framework {\methodname} to decode visual stimuli using only a few paired fMRI-image annotations from brain recordings. 
In Stage A, we employ an fMRI pre-training scheme with masked modeling to learn generalizable context knowledge from a large-scale unlabeled fMRI dataset. In Stage B, we use a latent diffusion model with double conditioning to generate plausible seen images from learned fMRI representations. We validated the decoding results of {\methodname} on multiple datasets and showed that our model generates more plausible and semantically similar images compared to previous methods, pushing the state-of-the-art a considerable step forward.

\clearpage
{\small
\bibliographystyle{ieee_fullname}
\bibliography{egbib}
}

\appendix\onecolumn
\section*{Appendix}
\renewcommand\thefigure{\thesection.\arabic{figure}} 
\renewcommand\thealgorithm{\thesection.\arabic{algorithm}} 
\renewcommand\thetable{\thesection.\arabic{table}} 
\setcounter{table}{0} 
\setcounter{figure}{0} 

\section{More Generation Samples}
\label{sec:allfigs-appendix}

All samples are generated at a resolution of $256\times256\times3$ with 250 PLMS~\cite{plms} steps. More samples can be found and generated in our code base. 

\begin{figure}[ht]
\centering
  \includegraphics[width=0.95\textwidth]{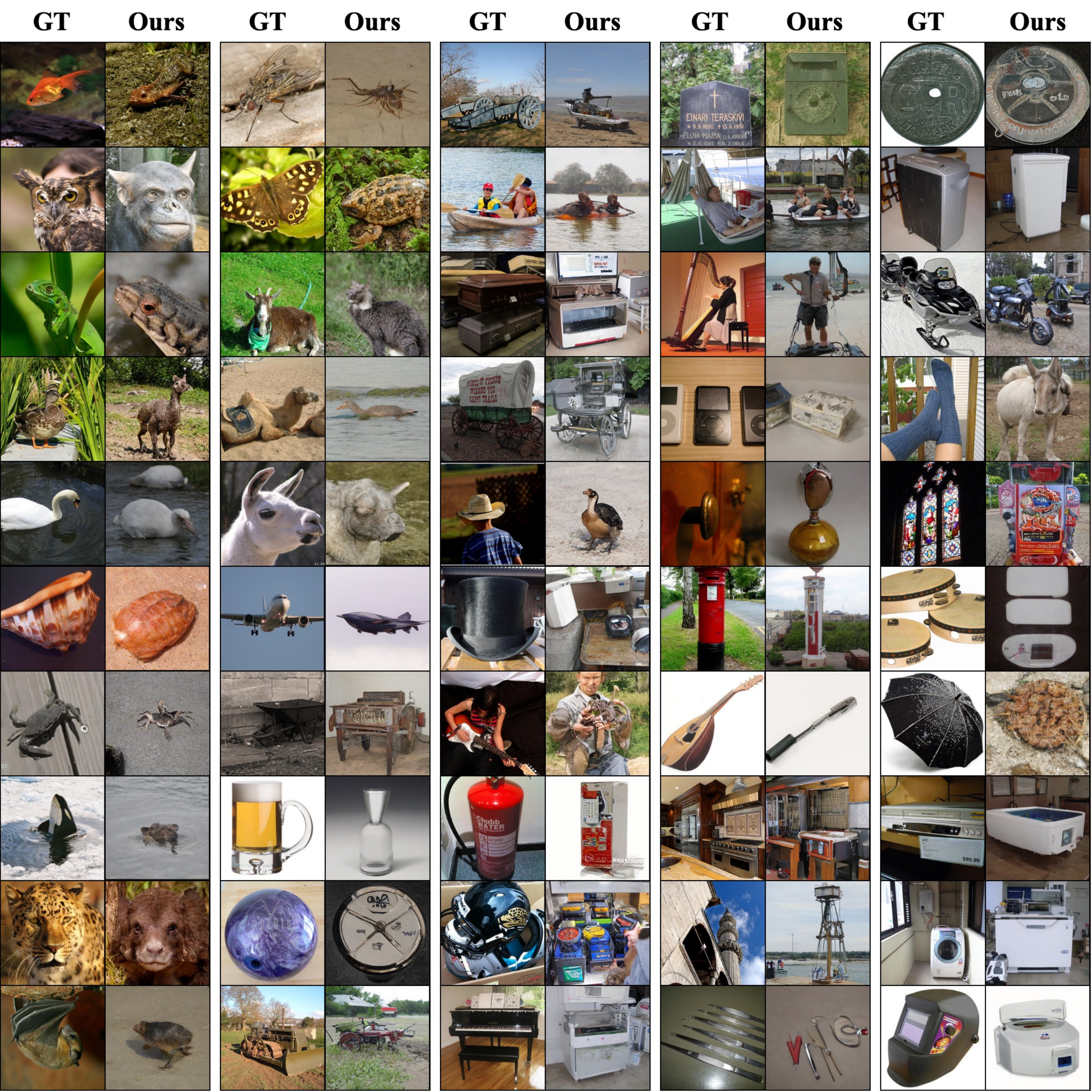}
  \caption{Full Samples for Subject 3 in GOD.}
     \label{fig:god_3}
\end{figure}

\begin{figure}[ht]
\centering
  \includegraphics[width=0.95\textwidth]{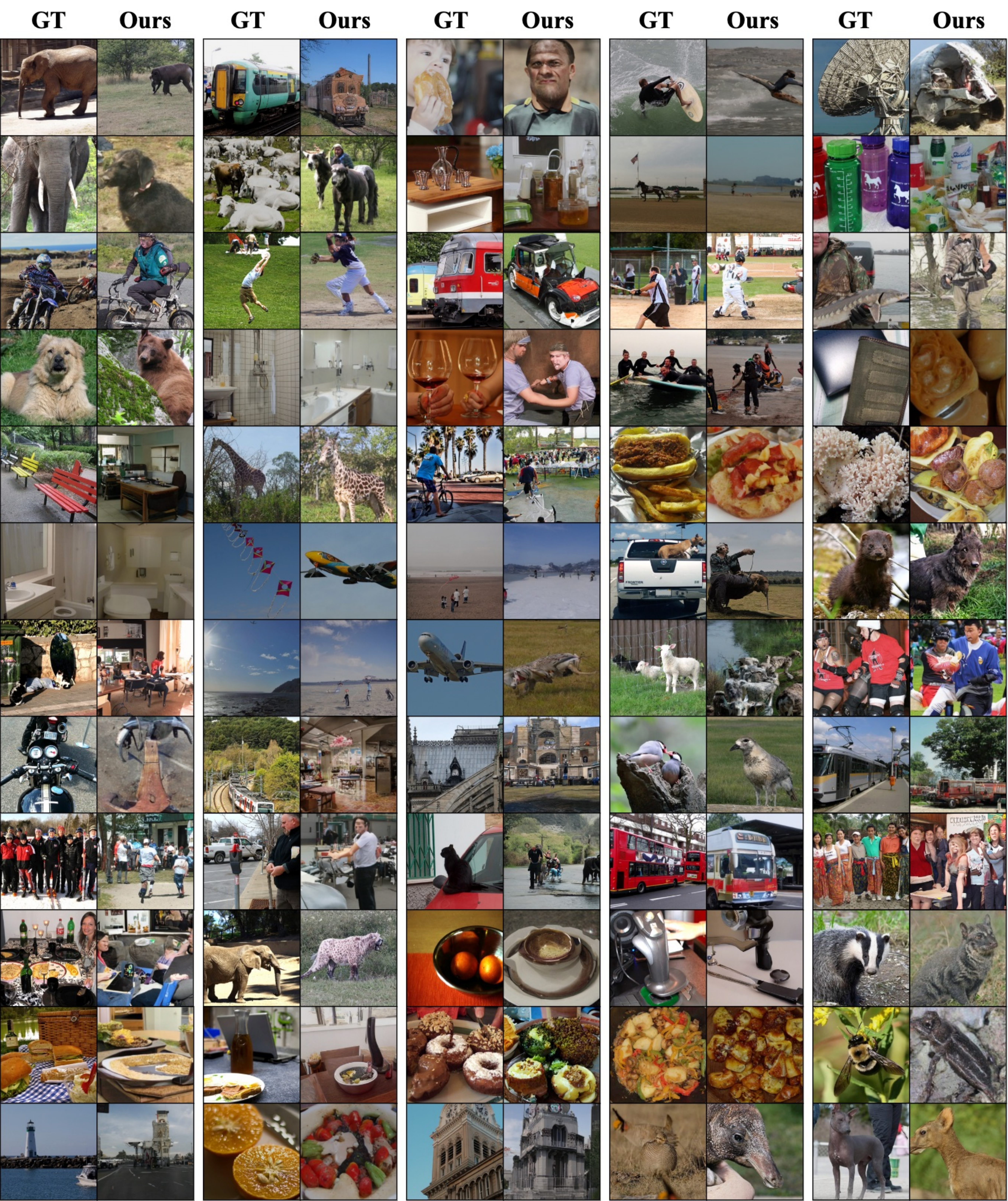}
  \caption{Full Samples for BOLD5000(Cont.).}
     \label{fig:bold5000_1}
\end{figure}

\begin{figure}[ht]
\centering
  \includegraphics[width=0.95\textwidth]{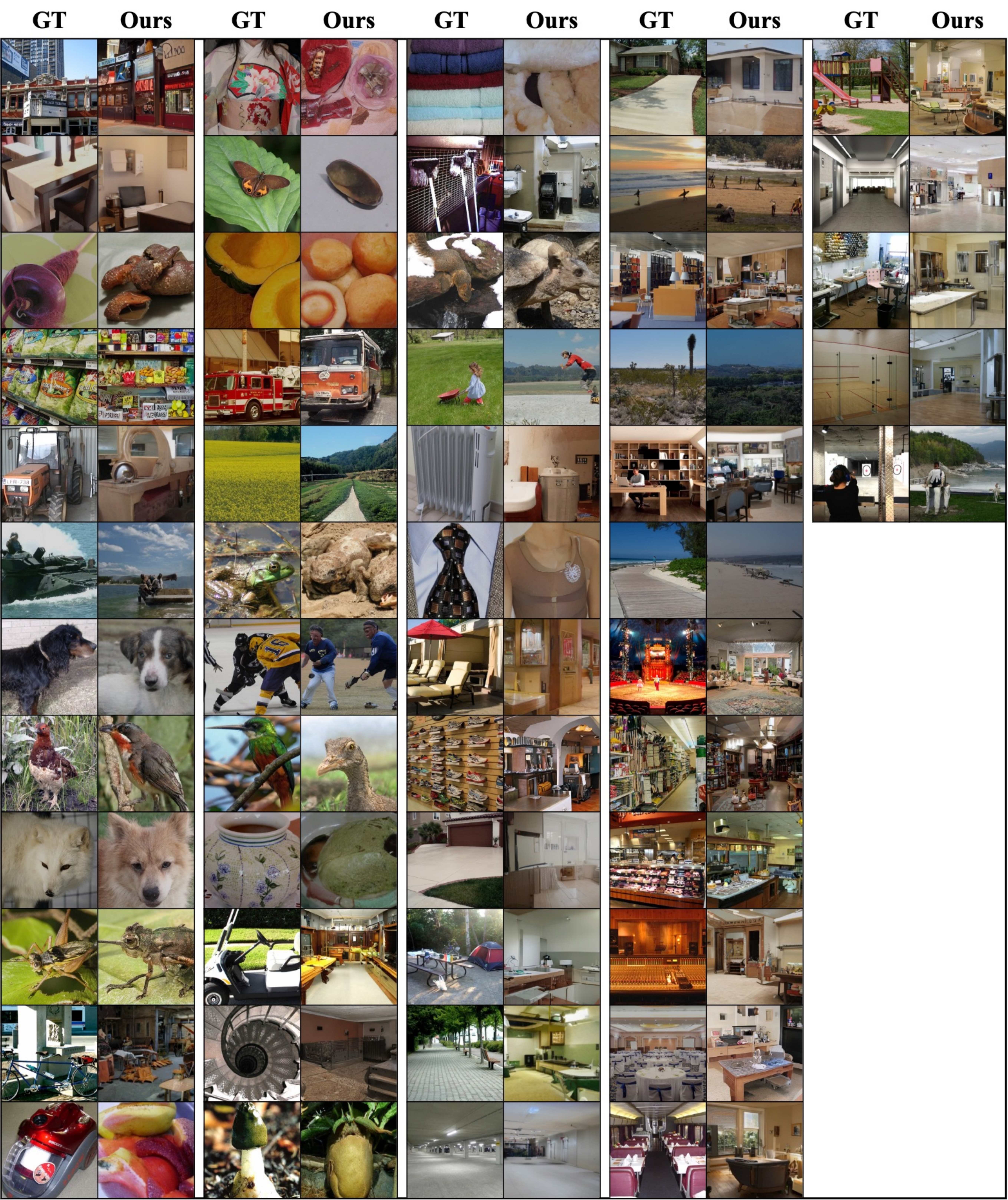}
  \caption{Full Samples for BOLD5000.}
     \label{fig:bold5000_2}
\end{figure}

\begin{figure}[ht]
\centering
  \begin{subfigure}[c]{0.2\textwidth}
         \centering
         \includegraphics[width=\textwidth]{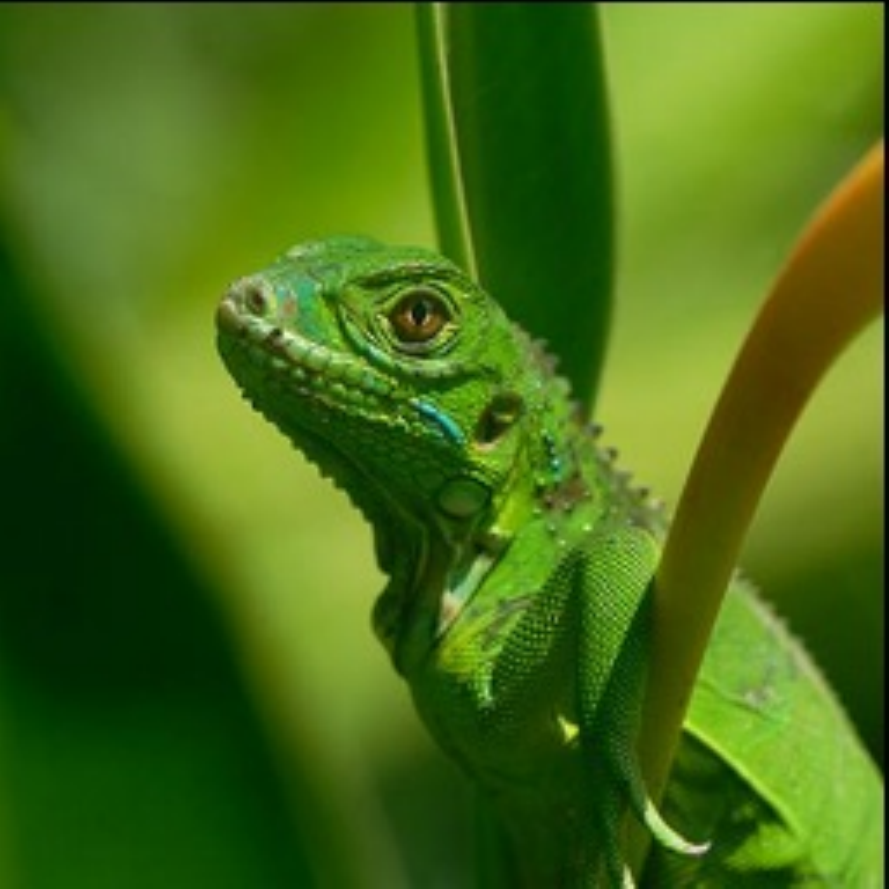}
         \caption{GT}
 \end{subfigure}
 \hfill
 \begin{subfigure}[c]{0.2\textwidth}
         \centering
         \includegraphics[width=\textwidth]{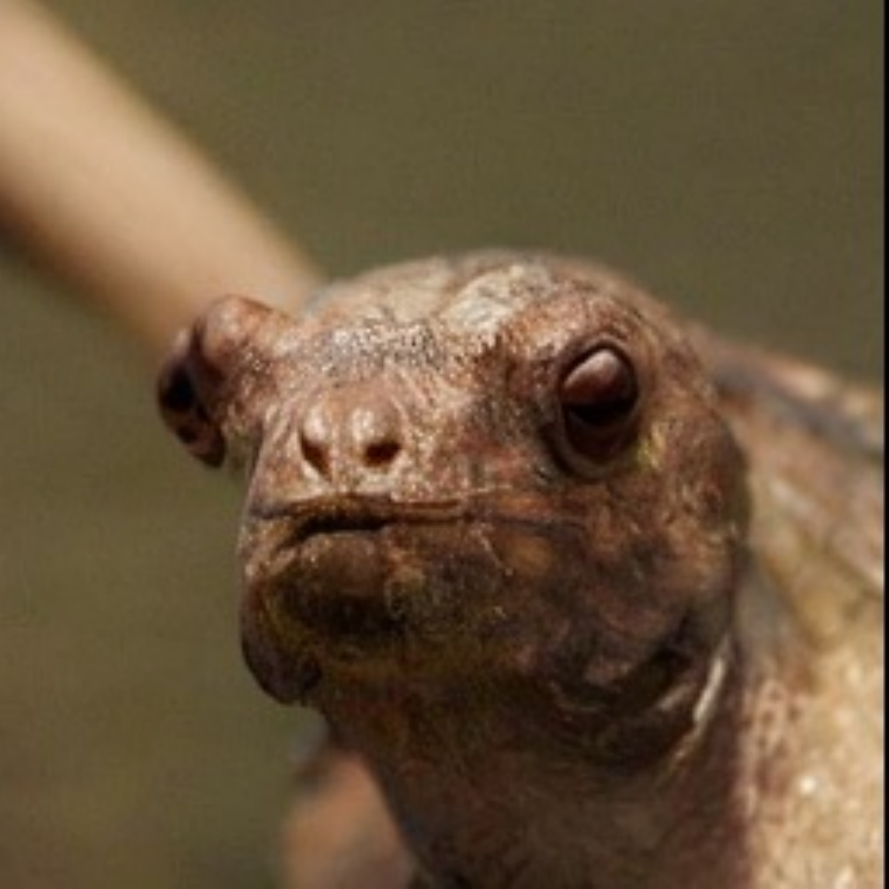}
         \caption{Full Model}
 \end{subfigure}
 \hfill
 \begin{subfigure}[c]{0.2\textwidth}
         \centering
         \includegraphics[width=\textwidth]{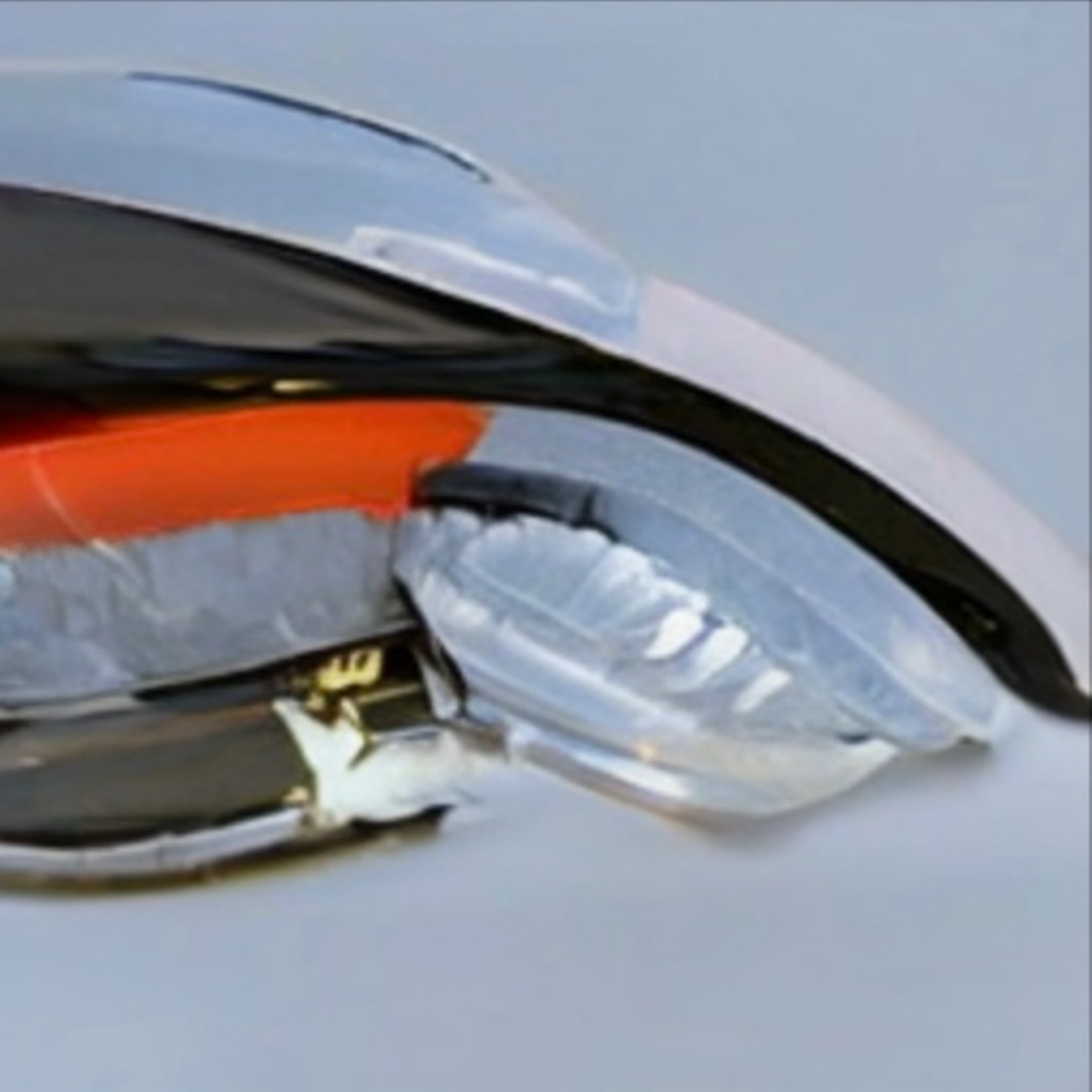}
         \caption{w/o SC-MBM Small}
 \end{subfigure}
 \hfill
 \begin{subfigure}[c]{0.2\textwidth}
         \centering
         \includegraphics[width=\textwidth]{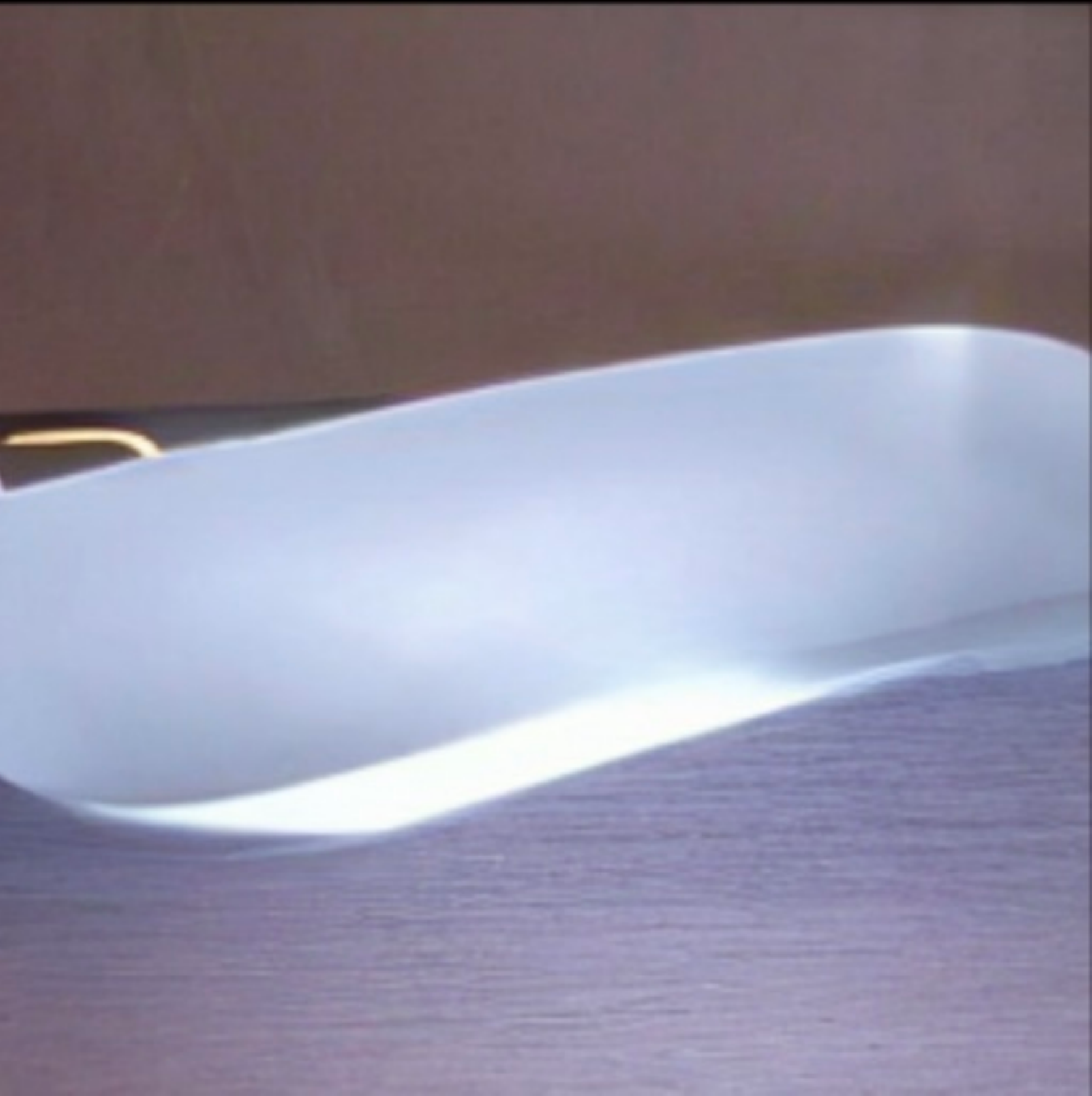}
         \caption{w/o SC-MBM Large}
 \end{subfigure}
 \hfill
 \begin{subfigure}[c]{0.2\textwidth}
         \centering
         \includegraphics[width=\textwidth]{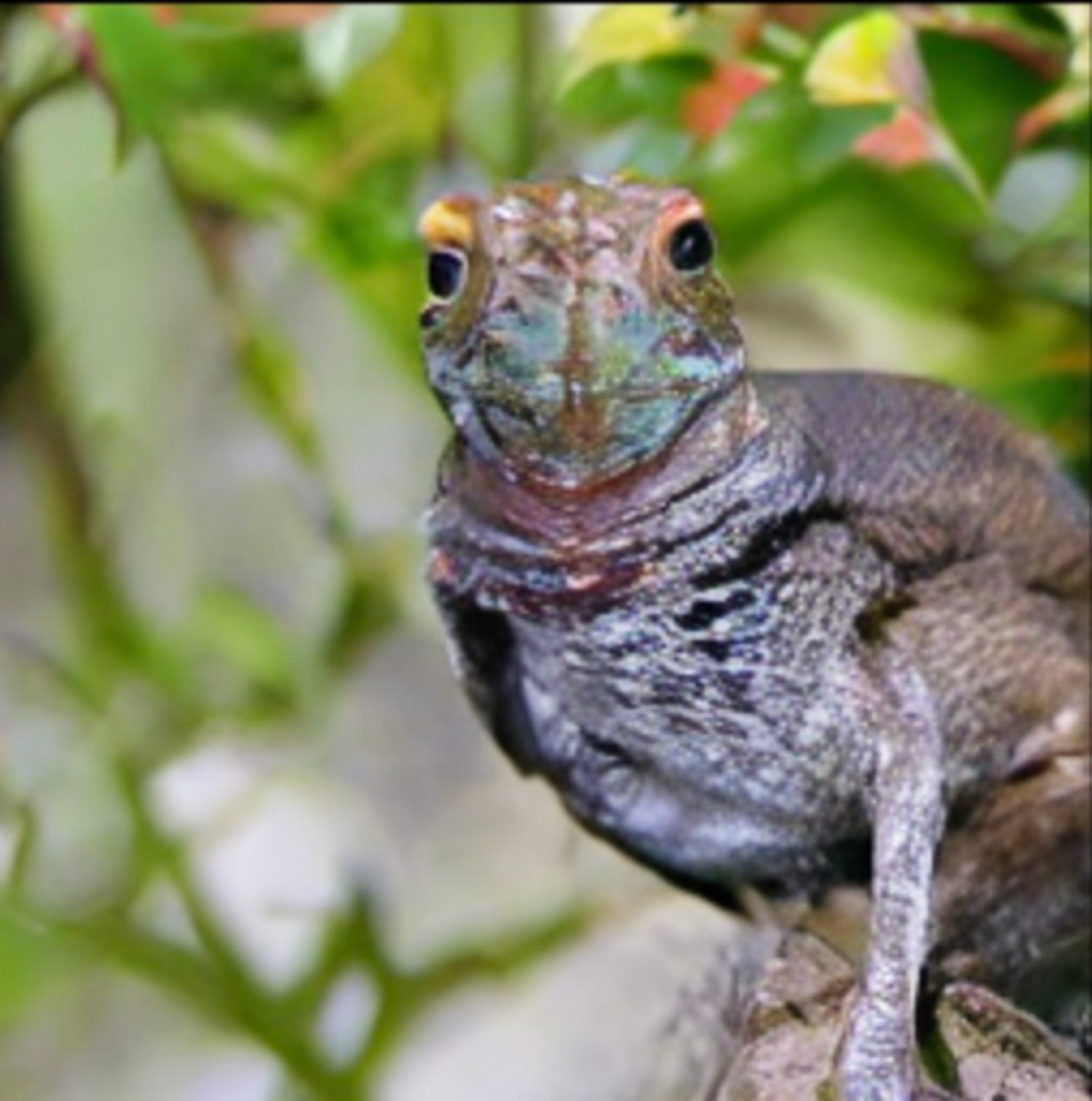}
         \caption{Optimize Encoder only}
 \end{subfigure}
 \hfill
 \begin{subfigure}[c]{0.2\textwidth}
         \centering
         \includegraphics[width=\textwidth]{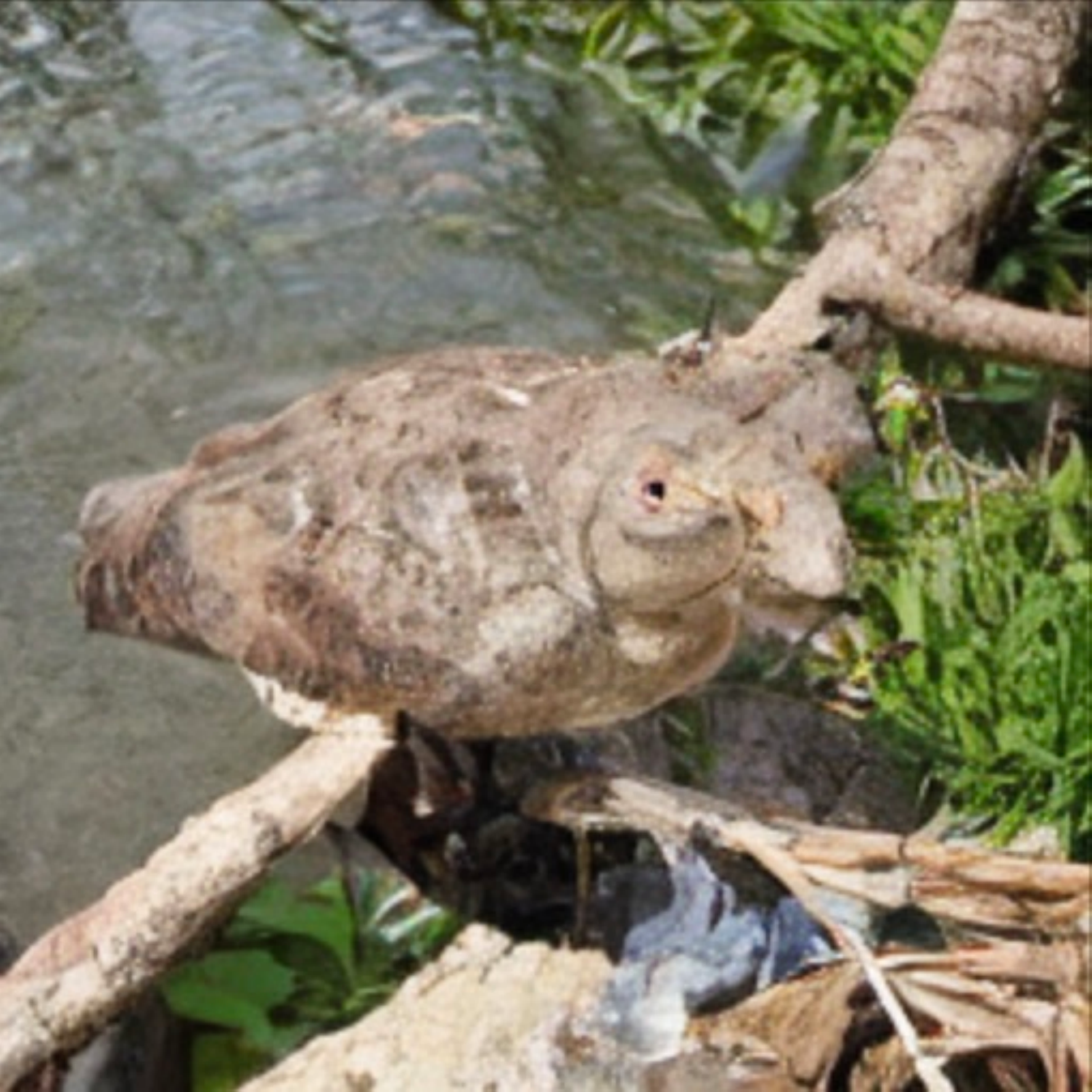}
         \caption{Single Conditioning}
 \end{subfigure}
 \hfill
 \begin{subfigure}[c]{0.2\textwidth}
         \centering
         \includegraphics[width=\textwidth]{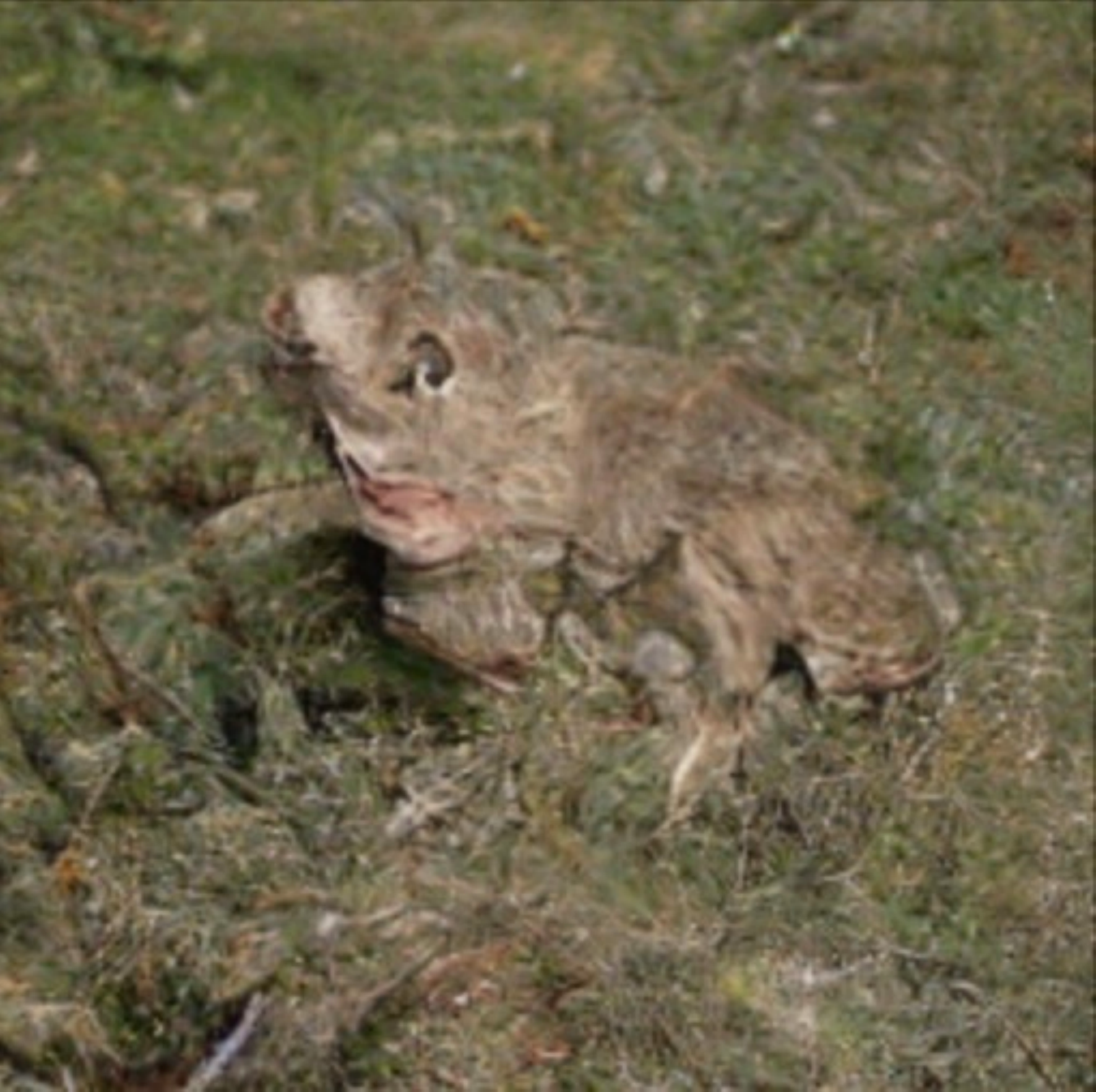}
         \caption{Text-to-Image Pre-trained LDM (LAION)}
 \end{subfigure}
 \hfill
 \begin{subfigure}[c]{0.2\textwidth}
         \centering
         \includegraphics[width=\textwidth]{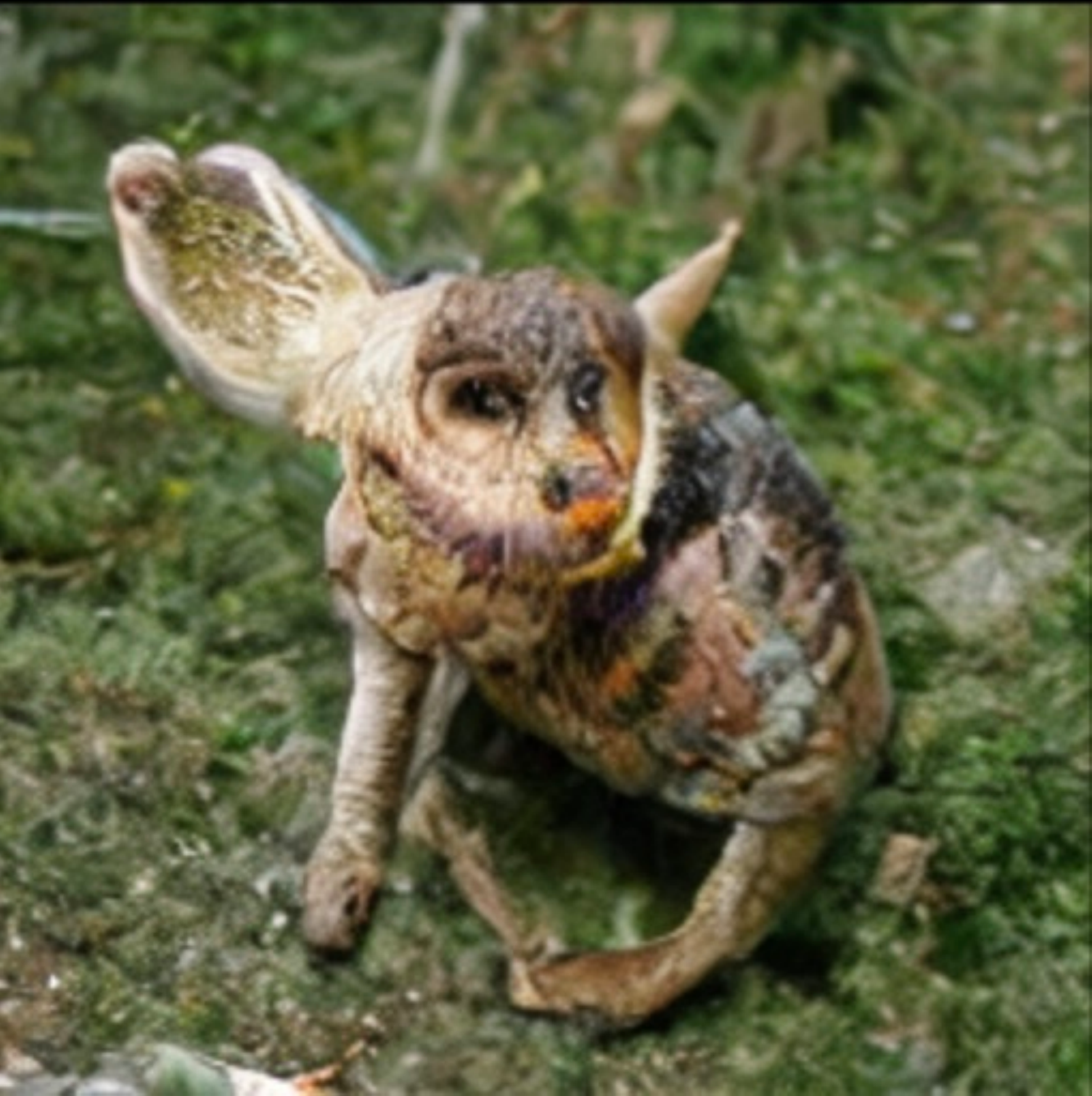}
         \caption{Layout-to-Image Pre-trained LDM (OpenImages)}
 \end{subfigure}
 
  \caption{Samples for Ablation Study. (a) Ground-truth stimulus. (b) Full model: SC-MBM pre-training; optimize the fMRI encoder and cross-attention heads; double-conditioned; Label-to-Image pre-trained LDM (ImageNet). (c) Model with small fMRI encoder without SC-MBM pre-training. (d) Model with the same fMRI encoder as the Full model without SC-MBM pre-training. (e) Optimize the fMRI encoder only, keep the cross-attention heads untouched. (f) Single conditioning. All other parameters are the same with the Full Model. The samples are obtained after finetuned for 500 epochs. See \cref{tab:MBM_test} and \cref{tab:ldm_design} for quantitative results of full test samples. }
     \label{fig:mbm-ablation-fig}
\end{figure}

\begin{figure}[ht]
  \centering
  \includegraphics[width=0.6\linewidth]{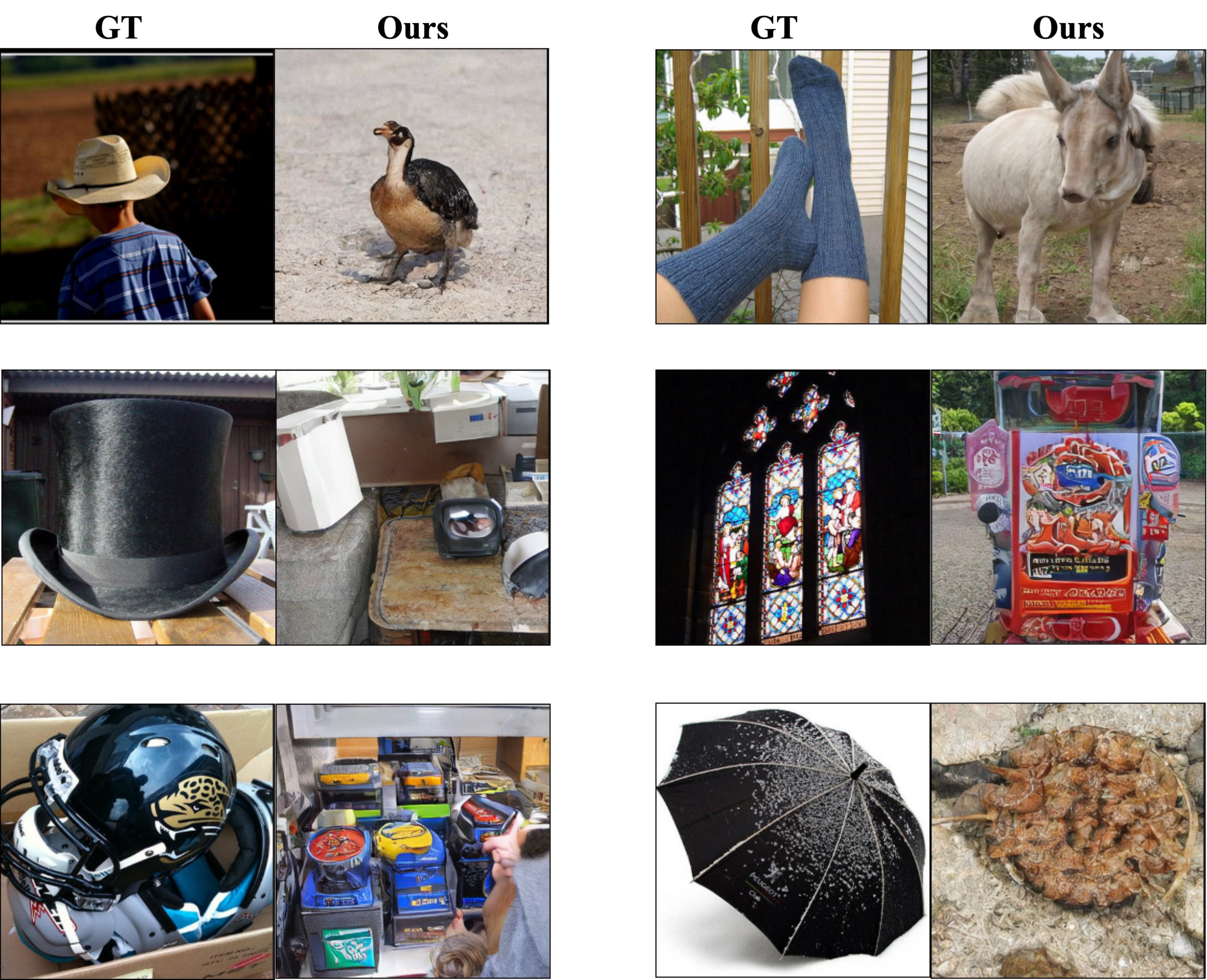}
   \caption{Typical Failure Cases of Our Method. As discussed in the main text, we assume the failure cases are related to two reasons. On one hand, the GOD training set and testing set have no overlapping classes. That is to say, the model could learn the geometric information from the training but cannot infer unseen classes in the testing set. On the other hand, subjects might have some other stimuli-unrelated thoughts, which could be captured by fMRI and decoded by our method}
   \label{fig:fail_cases}
\end{figure}

\clearpage

\section{Dataset and fMRI Preprocessing Details}
\label{sec:datasets} 
\textbf{Human Connectome Project (HCP) 1200 Subject Release~\cite{hcp}}: large-scale magnetic resonance imaging dataset used for pre-training. We utilized around $2000 \times 15$-min 3T resting state fMRI runs from 1091 subjects. The visual cortex (V1-V4) defined in \cite{hcproi} is used as the ROI, which gives approximately 4000 voxels.

\textbf{Generic Object Decoding Dataset (GOD)}~\cite{kam2017}: human fMRI scans with 1250 distinct images from ImageNet as a visual stimulus. 
During the fMRI scan, subjects were instructed to fixate on a cross located at the center of the presented images. 
This dataset consists of 1250 natural images from 200 distinct classes from ImageNet, where 1200 images are used for training. The remaining 50 images from classes not present in the training data are used for testing. 
Each image in the training set is presented once to the subject during the scan, while each image in the testing set is presented 35 times. 
Following the preprocessing in~\cite{kam2017}, the 35 repetitions are averaged for each image to create a higher SNR fMRI sample for testing. 
This dataset is widely used in brain image decoding~\cite{biggan2020,guy2019,guy2022,shen2019,shen2019gan,fang2020reconstructing}. We used the manually defined ROIs (V1-V4, FFA, LOC, HVC) from the functional localizer runs provided in \cite{kam2017}. 
Altogether, the ROIs have around 4500 voxels per subject, with some individual variance as shown in~\cref{fig:individual_diff}. 

\textbf{Brain, Object, Landscape Dataset (BOLD5000)~\cite{chang2019bold5000}}: human fMRI study with 5,254 fMRI-image pairs from 4,916 distinct natural images (including various objects and indoor/outdoor scenes) from Scene UNderstanding (SUN) \cite{xiao2010sun}, Common Objects in Context (COCO)\cite{lin2014microsoft} and ImageNet\cite{imagenet1k}. 
In this dataset, 4,803 images are presented once, and 113 images are repeated twice or three times. 
The repeated data are also averaged to construct the testing set as in the GOD dataset. 
To the best of our knowledge, this dataset is the first time is applied to an image reconstruction task. 
The author also provided manually annotated ROIs based on a functional localizer. 
As a result of different scanning resolutions and ROI definition methods, the number of voxels in the defined ROIs is approximately 1,500 for each subject. 
Nonetheless, we show in our results that the pre-trained encoder can be directly applied to this dataset despite this difference in ROI definition and size.   

\textbf{Pre-training dataset}
Our upstream pre-training dataset is comprised of fMRI recordings from HCP and GOD.
Following the processing step in target dataset\cite{kam2017}, we averaged every 8.64 seconds (\ie 12 time frames) of scans from HCP, which gives 130,000 fMRI time points.
Including the training and testing fMRI in GOD, we have a pure fMRI dataset of 136,000 samples for pre-training.
This pre-training dataset is, by far, the largest pre-training fMRI dataset used in this task.   

To handle the different voxel numbers, all fMRI data are first padded to the maximum length in a wrap-around manner and then padded to the boundary of the patch size.
Additionally, training fMRI is normalized to have zero mean and unit standard deviation. 
The testing samples are normalized with the mean and standard deviation from the training set.

\section{Results on Different Subjects}
The GOD consists of five different subjects, and the BOLD5000 consists of four subjects. The signal-to-noise ratio (SNR) is usually used to quantify the quality of a dataset. A higher SNR means better data quality.  As reported by their authors respectively, the BOLD5000 has a much higher SNR than GOD. Within the GOD, the SNR differs among subjects as shown in \cref{tab:subjects}, where Subject~3 has a significantly higher SNR than the others. A higher SNR leads to better performance in our experiments, which has also been shown in various literature. Other than possible noise introduced during the scan, the SNR is also related to the subjects' on-line processing or information processing ability. Subjects with better information processing ability (\ie better learners) will have a higher SNR during the scan under the same scanning conditions.    

\begin{table}[ht]
\centering
\begin{tabular}{|*{10}{c|}}
 \hline\hline
 Dataset & \multicolumn{5}{c|}{GOD} & \multicolumn{4}{c|}{BOLD5000} \\ 
 \hline
 Subject & Sub1 & Sub2 & Sub3 & Sub4 & Sub5 & CSI1 & CSI2 & CSI3 & CSI4 \\
 \hline
 Acc (\%) & 9.1 & 13.9 & 27.4 & 15.2 & 14.3 & 34.5 & 18.5 & 21.0 & 20.9  \\
 \hline
 FID & 2.2 & 1.6 & 1.7 & 2.7 & 2.4 & 1.2 & 1.9 & 1.4 & 1.3 \\
 \hline
 SNR     & 0.064\scriptsize{$\pm0.07$} & 0.061\scriptsize{$\pm0.05$} & 0.10\scriptsize{$\pm0.11$} & 0.092\scriptsize{$\pm0.1$} & 0.065\scriptsize{$\pm0.06$} & 4.65\scriptsize{$\pm0.2$} & 5.20\scriptsize{$\pm0.2$} & 5.55\scriptsize{$\pm0.35$} & 5.40\scriptsize{$\pm0.1$} \\
 \hline\hline
\end{tabular}
\caption{Full Results for All Subjects. The accuracy is obtained from the 1000-trials 50-way top-1 semantic classification test on the best-generated samples. SNR: signal-to-noise ratio. The voxel-wise mean SNR is obtained from \cite{kam2017} and \cite{chang2019bold5000} respectively.}
\label{tab:subjects}
\end{table}

\section{More Implementation Details}
\label{sec:hyperparam-appendix}

\subsection{Evaluation Metric Implementation}
This algorithm performs the N-trial, n-way top-1 semantic classification test. It measures the semantic accuracy of generated images. 
We describe our evaluation method in~\cref{alg:metric}, where the generated image and its corresponding ground-truth image are denoted by $x$ and $\hat{x}$ respectively, and $y$ is for the class label. 
This metric relies on a pre-trained ImageNet classifier to determine whether $x$ and $\hat{x}$ belong to the same class rather than using handcrafted features to represent each class. This method is thus reasonable and easily reproducible. 
We used a pre-trained ResNet as the classifier. 
We also showed that using other model based pre-trained classifiers will not change the result of this metric. 

\begin{algorithm}[H]
    \caption{N-Trials n-way Top-1 Accuracy Classification}\label{alg:metric}
    \begin{algorithmic}[1]
        \State \textbf{Input} pre-trained classifier $\mathcal{C}(\cdot)$,  image pair (Generated Image $x$, Corresponding GT Image $\hat{x})$
        \State \textbf{Output} success rate $r\in[0,1]$
        \For{$N$ trials} 
            \State $\hat{y} \leftarrow \mathcal{C}(\hat{x})$ get the ground-truth class
            \State $\{p_0,...,p_{999}\} \leftarrow \mathcal{C}(x)$ get the output probabilities
            \State $\{p_{\hat{y}}, p_{y_1}, ..., p_{y_{n-1}}\} \leftarrow$ pick $n$-1 random classes
            \State success if $\mathop{\arg\max}\limits_y\{p_{\hat{y}}, p_{y_1}, ..., p_{y_{n-1}}\} = \hat{y}$
        \EndFor
        \State $r=$ number of success / $N$
    \end{algorithmic}
\end{algorithm}

\subsection{SC-MBM Pre-training}
In Masked Image Modeling (MIM)~\cite{maeHe}, images are divided into patches which are sequentially transformed into embeddings to adapt to a transformer-based architecture~\cite{vit}.
Following this practice, we divided fMRI voxels into patches which will be subsequently transformed into embeddings using a one-dimensional convolutional layer with a stride of the patch size. 

A patch size of 16 and an embedding dimension of 1024 were used as the Full model. 
Notice that our embedding size to patch size ratio is much larger than that of MIM. 
For example in \cite{maeHe}, the authors used a patch size of 16 and embedding dimension 768, which gave an embedding to patch dimension ratio: $768/(16\cdot16\cdot3)=1$, compared to ours: $1024/(16)=64$. 
This design largely expands the representation dimension of fMRI data, significantly boosting the information capacity of the fMRI representations. 
This design is justified by both considering the dimension gap between fMRI and natural images, as well as the hypothesis of sparse coding in the visual encoding process.

Following~\cite{simmim}, we adopt an asymmetric architecture where the decoder is much smaller than the encoder. 
Before feeding patch embeddings to the encoder, a random portion is masked. 
We used a large mask ratio similar to the mask ratio used in MIM due to the similarity in information density between fMRI data and images. 
We additionally embed mask tokens and include positional embeddings along with the patch encodings at the end of the encoder and transform them into the decoder's embedding space via a linear projector. 
On the other hand, our decoder aims to recover the masked patches with the voxel value as the prediction target. 

To train the data-hungry model like the ViT, we also applied random sparsification (RS) for data augmentation, where $20\%$ of voxels in each fMRI were randomly selected and set to zero.

\begin{figure}[ht]
\centering
  \includegraphics[width=0.9\textwidth]{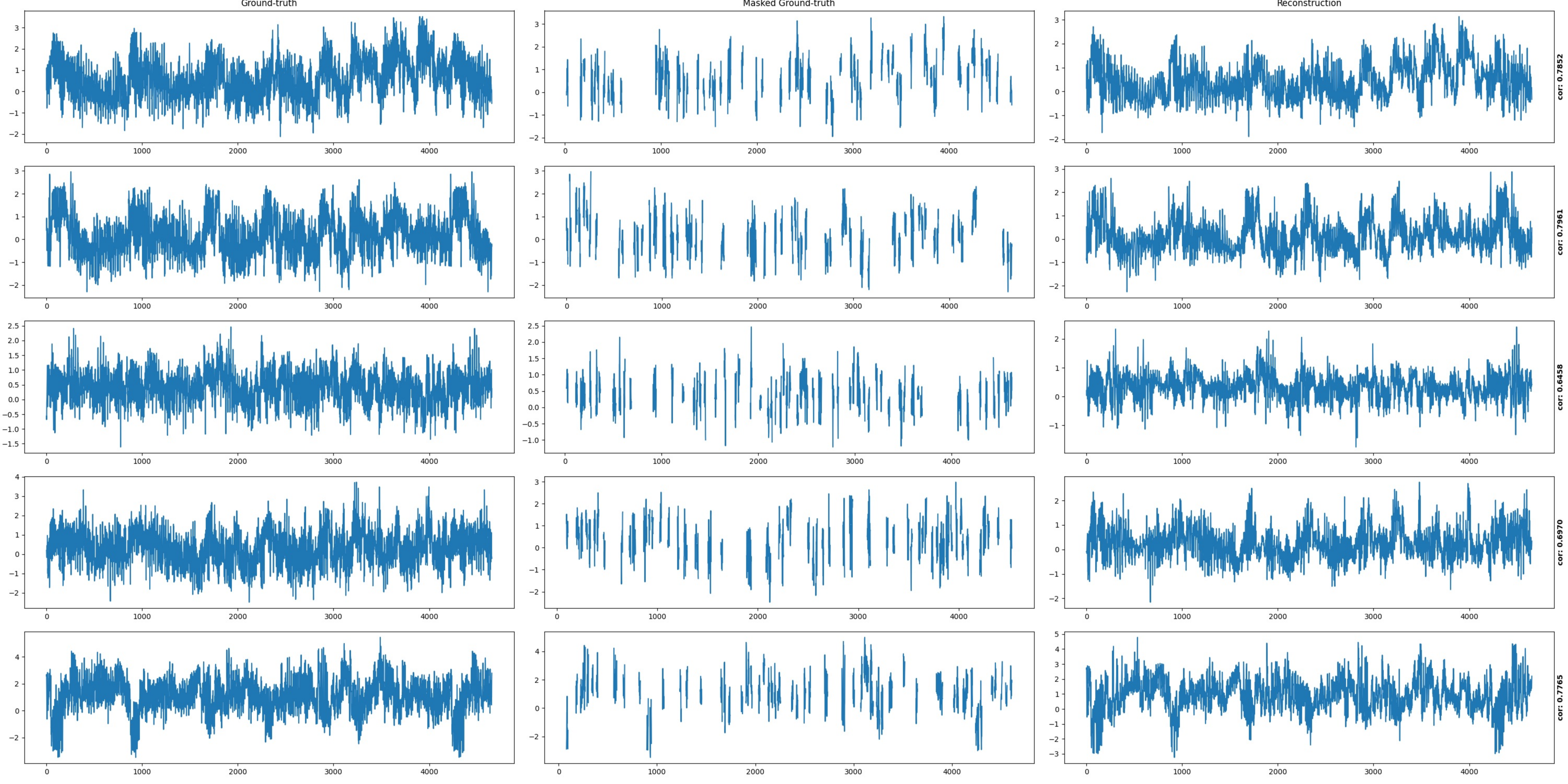}
  \caption{Examples of masked brain prediction. First column: original fMRI data (Visual Cortex) flattened; Second column: masked fMRI; Third column: data recovered from SC-MBM decoder. Mask ratio: 0.75. The correlations between the original and recovered fMRI are also shown.}
     \label{fig:mbm_mask_exp}
\end{figure}

Hyperparameters used in the SC-MBM pre-training stage are listed in \cref{tab:mbm_para}. All other unlisted parameters are set to their defaults. The SC-MBM pre-training is performed on 8 RTX3090ti GPUs until the model converges. Examples of masked brain prediction are shown in \cref{fig:mbm_mask_exp}.

\begin{table}[ht]
\centering
\begin{tabular}{|lc|lc|lc|lc|}
 \hline\hline
 parameter&value&parameter&value&parameter&value&parameter&value \\
 \hline \hline
 patch size & 16 & encoder depth & 24 & decoder embed dim & 512 & clip gradient & 0.8\\ 
 embedding dim & 1024 & encoder heads & 16 & max learning rate & 2.5e-4 & weight decay & 0.05\\ 
 mask ratio & 0.75 & decoder depth & 8 & warm-up epochs & 40 & batch size & 500\\ 
 mlp ratio & 1.0 & decoder heads & 16  & max epochs & 500 & optimizer & AdamW~\cite{adamw} \\
\hline\hline
\end{tabular}
\caption{Hyperparameters used in the Full model for SC-MBM Pre-training. }
\label{tab:mbm_para}
\end{table}

\subsection{DC-LDM Finetuning}
The finetuning is performed by jointly optimizing the fMRI encoder and cross-attention heads in the LDM using the training set. Specifically, for an fMRI-image pair, the image will be encoded into the latent space via a VQ encoder, which will be subsequently used as an objective to train the fMRI encoder and cross-attention heads. In the forward pass, fMRI data is passed through the encoder, producing a patchified enlarged representation. Then this representation is projected into an intermediate space with a channel size of $M$. This intermediate representation will be used as the key and value in cross-attention modules in the UNet and will also be added to the time embedding used in the UNet. The UNet tries to denoise a Gaussian noise with the fMRI representation as a condition, mimicking the reverse transitions through a Markov Chain. L2 loss is used in training. During the training, only the fMRI encoder and cross-attention modules in the LDM are optimized. Other parts are kept intact. 

Operating in the image latent space, the computations needed for DC-LDM finetuning are small. All finetunings in our experiments are performed with a single RTX3090ti GPU for 500 epochs. The detailed hyperparameters are shown in \cref{tab:ldm_para}. All other unlisted parameters are set to their
defaults. Please see \cite{ldm2022} for the detailed model architecture of the LDM.
\begin{table}[ht]
\centering
\begin{tabular}{|lc|lc|lc|lc|}
 \hline\hline
 parameter & value & parameter & value & parameter & value & parameter & value \\
 \hline \hline
 batch size & 5 & diffusion steps & 1000 & image latent dim & $64\times64\times3$ & learning rate & 5.3e-5\\ 
 \hline
 image resolution & $256\times256\times3$ & optimizer & AdamW & pre-trained type & Label-to-Image & $M$ & 77 \\
\hline\hline
\end{tabular}
\caption{Hyperparameters used in the Full model for DC-LDM Finetuning. }
\label{tab:ldm_para}
\end{table}

\section{Other Ablation Studies}

\label{sec:ablations-appendix}

\paragraph{Patch Sizes}
In \cite{vit}, an image is divided into sixteen $16\times 16$ patches which can be considered 16 words. Analogous to the fMRI data, the more words are used to describe the data, the higher accuracy of the resulting representation will be. Therefore, smaller patches will lead to better results if the number of voxels remains unchanged. This claim is justified by \cref{tab:other_abl}. A continuous decrease in accuracy can be observed when the patch size is increased from 16 to 64. However, the minimal patch size applicable is constrained by available memory, as the number of patches increases drastically with smaller patch size.  

\paragraph{Encoder Depth}
The fMRI encoder depth is set to 24 in our Full model similar to the ViT-Large~\cite{vit}. However, different depths lead to a different number of parameters and encoding capabilities. Usually, a deeper model is appreciated, but it comes with the need for more training samples as well. Therefore, considering the limited data, we explore whether a smaller model would have better results. To maintain an asymmetrical architecture, the depth of the SC-MBM decoder is kept at half of the encoder's depth. A deeper fMRI encoder (as deep as 24 transformer blocks) gives the best result as shown in \cref{tab:other_abl}.

\paragraph{Mask Stragtey}
In \cite{maeHe}, different masking strategies are tested for images, and the authors conclude that random masking is the best strategy for images. We explore in our ablation if it is the case for fMRI learning. For images, there are different strategies such as center masking and grid masking due to the geometric correlations among pixels in an image. For fMRI data, brain activities are reflected by the connectivity among groups of voxels (functional networks). Seven networks in the visual cortex are used in our study (\ie V1-V4, FFA, PPA, and LOC), in which the V1, the primary visual cortex, consists of the most voxels and is the first stage of visual processing. Therefore we design a focus masking strategy similar to the center masking in images. In the center masking, pixels at the center of an image will be masked the most because the center of an image usually contains the richest information. Learning to recover the center potentially is beneficial to learning the underlying semantics of an image. Similar to the center masking, our focus masking in fMRI masks more patches in the V1 region than in other regions. However, in our experiments, the focus masking does not outperform the random masking strategy as shown in \cref{tab:other_abl}.

\paragraph{Pretext Tasks}
As discussed, since fMRI voxels are correlated reflecting the underlying brain activities, masked modeling is a suitable learner for fMRI representations. With SC-MBM as a pretext task, self-supervised learning can be performed in a large unpaired fMRI dataset. On the other hand, considering a small part of paired fMRI are available in the training set. Therefore, it is intriguing if we can use the paired information in fMRI for the pre-training as well. So we include the image feature as another pretext task together with the masked modeling to guide the context learning. Specifically, the training set will be divided into two parts: the part with paired images; and the part without paired images. To construct a mini-batch, we randomly sample fMRI from these two parts. In this design, another decoder is added to decode image features. The image features extracted from the second layer of a pre-trained VGG will be used as a target for this decoder. During training, the image feature reconstruction loss will be added to the MBM loss with a regularization term. However, adding the image guidance does not outperform the MBM only pre-training as shown in \cref{tab:other_abl}.

\paragraph{Unequal Length Handle}
Due to individual variability, even in the same dataset, the voxel numbers of individuals are different. We need to handle this unequal length to include different subjects in the pre-training. The two most intuitive ways are considered: pad to the maximum length with a constant; cut to the minimum length. Besides padding with a constant, we pad the data in a wrap-around manner. From \cref{tab:other_abl} we can see that cutting the data gives the worst performance and wrap-around padding gives the best performance. 

\paragraph{Crop Ratio}
In the finetuning, images are randomly center-cropped for augmentations. We tested different crop ratios, \ie from 0 to 0.4. It is found that a crop ratio of 0.2 gives the best performance. Random cropping is an efficient augmentation in our task because the subjects' perceptions may be focused on different parts of the figure, even though they were instructed to fixate at the center of the image. 

\begin{table}[ht]
    \begin{subtable}{.3\linewidth}
        \centering
        \begin{tabular}{c|ccc}
             Patch size & \cellcolor[gray]{0.8}16 & 32 & 64\\
             \hline
             Acc (\%)   &  \cellcolor[gray]{0.8}23.9 & 18.2 & 16.4 \\
        \end{tabular}
        \caption{Patch Size}%
    \end{subtable}
    \hfill%
    \begin{subtable}{.3\linewidth}
        \centering
        \resizebox{\linewidth}{!}{%
        \begin{tabular}{c|ccc}
             Encoder depth & \cellcolor[gray]{0.8}24 & 8 & 2\\
             \hline
             Acc (\%) &  \cellcolor[gray]{0.8}23.9 & 14.8 & 13.6 \\
        \end{tabular}}%
        \caption{fMRI Encoder Depth}%
    \end{subtable}
    \hfill%
    \begin{subtable}{0.3\linewidth}
        \centering
        
        \begin{tabular}{c|cc}
             Strategy & \cellcolor[gray]{0.8}random & focus\\
             \hline
             Acc (\%) &  \cellcolor[gray]{0.8}23.9 & 16.3 \\
        \end{tabular}
        \caption{Mask Strategy}%
    \end{subtable}
    \vspace{3pt}
    \begin{subtable}{0.3\linewidth}
        \centering
        \resizebox{\linewidth}{!}{%
        \begin{tabular}{c|cc}
             Task & \cellcolor[gray]{0.8}SC-MBM & SC-MBM+image\\
             \hline
             Acc (\%) &  \cellcolor[gray]{0.8}23.9 & 16.1 \\
        \end{tabular}}%
        \caption{Pretext Tasks}
    \end{subtable}
    \hfill%
    \begin{subtable}{0.3\linewidth}
        \centering
        \begin{tabular}{c|ccc}
             Strategy & \cellcolor[gray]{0.8}wrap & constant & cut\\
             \hline
             Acc (\%) &  \cellcolor[gray]{0.8}23.9 & 19.6 & 14.8\\
        \end{tabular}
        \caption{Unequal Length Handle}
    \end{subtable}
    \hfill%
    \begin{subtable}{0.3\linewidth}
        \centering
        \resizebox{\linewidth}{!}{%
        \begin{tabular}{c|cccc}
             Ratio & 0 & 0.1 &\cellcolor[gray]{0.8}0.2 & 0.4\\
             \hline
             Acc (\%) & 14.9 & 17.9 & \cellcolor[gray]{0.8}23.9 & 15.2 \\
        \end{tabular}}
        \caption{Crop Ratio}
    \end{subtable}
    
    \caption{Other Design Ablations. The 1000-trial 50-way top-1 semantic classification accuracy is reported. All ablations are pre-trained for 500 epochs and then finetuned on GOD for another 500 epochs. Settings used in the Full model are colored in gray. }%
    \label{tab:other_abl}
\end{table}

\section{Extra Notes on Sparse-Coded Masked Brain Modeling}
In our design, we use a large embedding-size-to-patch-size ratio to increase the information capacity of fMRI representations, which mimics the sparse coding mechanism underlying the encoding procedure of the visual cortex. Here, we provide a formal definition of the information capacity and explain the connection with the sparse coding  mechanism. 

\begin{definition}
(Data Representation) For a piece of data given by a one-dimensional code vector $x\in\mathbb{R}^L$, let $f$ be an injective function that maps $x$ from the data domain to a representation domain, namely $f(x)=y$, where $y\in\mathbb{R}^{\tilde{L}}$ is a representation of $x$.
\end{definition}

\begin{definition}
(Information Capacity) For a random variable $X$, the Shannon entropy of $X$ is upper bounded by its cardinality, which is given by $H(X) \leq \log(|\mathcal{X}|)$. We define $\log(|\mathcal{X}|)$ as the information capacity of random variable $X$.
\label{def:info}
\end{definition}

The inequality in \cref{def:info} can be easily proved with Jensen's inequality regardless of the distribution of $X$. 
Obviously, for a representation $Y$, if the dimension of $Y$ is larger, the representation space will be larger. Hence, $Y$ will have a larger information capacity. To measure the change of information capacity after the representation mapping, we can simply divide the representation dimension by the data dimension, namely, $R=\tilde{L}/L$. In the context of masked modeling, we refer to $R$ as the embedding-size-to-patch-size ratio. 
The essence of sparse coding is to use sets of over-complete bases to efficiently represent data~\cite{sparse}. Analogous to this over-completeness, we use a representation space that is much larger than the data space, namely higher $R$, to learn the representations of the fMRI. Data locality is included in the representations by dividing the fMRI time series into patches and transforming patches into embeddings. 

\section{Pixel-level metrics}
 We also performed the pixel-level metrics (MSE \& LPIPS) for additional evaluation (Table below). Semantic-oriented methods, Ozelik\cite{icgan2022} and our approach outperformed the others in semantic metrics but not in pixel-level metrics. Pixel and semantic-level decodings recover visual stimuli from two perspectives, where the \textbf{trade-off between fidelity and meaningfulness} needs to be considered.
In this work, we \textbf{prioritize the recovery of visual semantics} in fMRI, which is crucial for understanding the complex mechanism of human perception.

\begin{table}[h]
\centering
\begin{tabular}{|c|c|c|}
\hline
Method & MSE$\downarrow$ & LPIPS$\downarrow$\\
\hline
Ours & 101 & 0.69 \\
\hline
Ozelik\cite{icgan2022} & 102 & 0.69  \\
Gaziv\cite{guy2022} & \textbf{99} & \textbf{0.68}\\
Beliy\cite{guy2019} & 105 & 0.81  \\
\hline
\end{tabular}
\caption{Comparison of Pixel-level Metrics Using MSE and LPIPS Benchmarks.}
\label{table:pixel-level-result}
\end{table}

\section{2-way and 5-way metrics}
 We also performed 2-way and 5-way metrics for additional evaluation (Table below). Our approach outperformed the others in both these two metrics. 

\begin{table}[h]
\centering
\begin{tabular}{|c|c|c|}
\hline
Method & 2-way$\uparrow$ & 5-way$\uparrow$\\
\hline
Ours & \textbf{0.86 }&\textbf{ 0.63} \\
\hline
Ozelik\cite{icgan2022} & 0.84 & 0.61  \\
Gaziv\cite{guy2022} & 0.71 & 0.39\\
Beliy\cite{guy2019} & 0.56 & 0.24  \\
\hline
\end{tabular}
\caption{Comparison of 2-way and 5-way Classification Metrics.}
\label{table:2-way-5-way-result}
\end{table}

\end{document}